\documentclass{article}
\usepackage[left=3.0cm, right=3.0cm]{geometry}
\usepackage{natbib}
\usepackage{times}
\usepackage[utf8]{inputenc} 
\usepackage[T1]{fontenc}    
\usepackage{hyperref}       
\usepackage{url}            
\usepackage{booktabs}       
\usepackage{amsfonts}       
\usepackage{nicefrac}       
\usepackage{microtype}      
\usepackage{authblk}
\usepackage{enumerate}
\usepackage{array}
\newcolumntype{M}[1]{>{\centering\arraybackslash}m{#1}}
\usepackage[tableposition=top]{caption}
\DeclareCaptionLabelFormat{andtable}{#1~#2  \&  \tablename~\thetable}
\usepackage{xr}
\usepackage{subfigure}
\usepackage{abstract}
\usepackage[toc, page]{appendix}

\title{\textbf{A Bayesian model for sparse graphs with flexible degree distribution and overlapping community structure}}

\author[1,4]{Juho Lee\thanks{Corresponding author, \texttt{juho.lee@stats.ox.ac.uk}}}
\author[2]{Lancelot F. James}
\author[3]{Seungjin Choi}
\author[1]{Fran\c{c}ois Caron}
\affil[1]{University of Oxford}
\affil[2]{The Hong Kong University of Science and Technology}
\affil[3]{Pohang University of Science and Technology}
\affil[4]{AITRICS}

\usepackage{amssymb, amsmath, amsthm}
\usepackage{mathtools}
\usepackage{bbm}
\usepackage{dsfont}
\usepackage[capitalize,nameinlink]{cleveref}

\newcommand{\bx}{\mathbf{x}}

\newcommand{\calL}{{\mathcal{L}}}

\newcommand{\calN}{{\mathcal{N}}}

\newcommand{\calS}{{\mathcal{S}}}

\newcommand{\calV}{{\mathcal{V}}}

\newcommand{\bbE}{\mathbb{E}}

\newtheorem{thm}{Theorem}[section]
\newtheorem{cor}{Corollary}[section]
\newtheorem{defn}{Definition}[section]

\newtheorem{prop}{Proposition}[section]

\newcommand{\dee}{\mathrm{d}}

\newcommand{\iidsim}{\overset{\mathrm{i.i.d.}}{\sim}}
\newcommand{\indicator}[1]{\mathds{1}_{\{#1\}}}
\newcommand{\defas}{\vcentcolon=}  

\def\[#1\]{\begin{align}#1\end{align}}

\newcommand{\pto}{\overset{p}{\to}}

\newcommand{\appropto}{\mathrel{\vcenter{
  \offinterlineskip\halign{\hfil$##$\cr
    \propto\cr\noalign{\kern2pt}\sim\cr\noalign{\kern-2pt}}}}}

\begin{document}

\maketitle

\renewcommand{\abstractnamefont}{\normalfont\large\bfseries}
\begin{abstract}
We consider a non-projective class of inhomogeneous random graph models with interpretable parameters and a number of interesting asymptotic properties. Using the results of \cite{Bollobas2007}, we show that i) the class of models is sparse and ii) depending on the choice of the parameters, the model is either scale-free, with power-law exponent greater than 2, or with an asymptotic degree distribution which is power-law with exponential cut-off. We propose an extension of the model that can accommodate an overlapping community structure. Scalable posterior inference can be performed due to the specific choice of the link probability. We present experiments on five different real-world networks with up to 100,000 nodes and edges, showing that the model can provide a good fit to the degree distribution and recovers well the latent community structure.
\end{abstract}

\section{Introduction}
\label{sec:introduction}
Simple graphs are composed of a set of vertices with undirected connections between them. The graph may represent a set of friendship relationships between individuals, a physical infrastructure network, or a protein-protein interaction network. Defining flexible and realistic statistical graph models is of great importance in order to perform link prediction or for uncovering interpretable latent structure, and has been the subject of a large body of work in recent years, see e.g.~\citep{Newman2009,Kolaczyk2009,Goldenberg2010}.

Our objective is to develop a class of models with interpretable parameters and realistic asymptotic properties. Of particular interest for this paper are the notions of sparsity and scale-freeness. A sequence of graphs is said to be sparse if the number of edges scales subquadratically with the number of nodes. The degree of a node is the number of connections of that node. The sequence of graphs is said to be scale-free if the proportion of nodes of degree $k$ is approximately $k^{-\eta}$ when the number $n$ of nodes is large, where the exponent $\eta$ is greater than 1. That is, for large $n$, the degree distribution behaves like a power-law. These notions of sparsity and scale-freeness have received a lot of attention in the network literature in the past years~\citep{Barabasi1999,Newman2009,Orbanz2015,Barabasi2016,Caron2017}; some authors argued that they are desirable properties of random graph models, and that many networks exhibit this scale-free behavior, usually with an exponent $\eta>2$. Other authors have recently challenged the scale-free assumption, showing that a power-law distribution with exponential cut-off provides a good fit to many real-world networks~\citep{Newman2009,Broido2018}, see the appendix for more discussion about testing for network scale-freeness. Besides these global asymptotic properties, we are also interested in capturing some latent structure in graphs. Individuals may belong to some latent communities, and their level of affiliation to the community defines the probability that two nodes connect.

We propose a class of sparse graph models with overlapping community structure and well-specified asymptotic degree distributions. The graph can either be scale-free with exponent $\eta>2$, or non-scale-free, with asymptotic degree distribution being a power-law distribution with exponential cut-off. The construction builds on inhomogeneous random graphs, a class of models exhibiting degree heterogeneity. This class of models has been studied extensively in the applied probability literature~\citep{Aldous1997,Chung2002,Bollobas2007,vanderHofstad2016}, but has been left unexplored for the statistical analysis of real-world networks. In \cref{sec:sparsedef} we provide a formal description of sparsity and scale-freeness for sequences of graphs. In \cref{sec:rank1} we describe the rank-1 inhomogeneous random graphs, and present their sparsity property and asymptotic degree distribution. The model is then extended in \cref{sec:rankc} in order to accommodate a latent community structure. Posterior inference is discussed in \cref{sec:inference}. In \cref{sec:discussion} we discuss the relative merits and drawbacks of our approach compared to other random graph models. \cref{sec:experiments} provides an illustration of the approach on several real-world networks, showing that the model can provide a good fit to the empirical degree distribution and recover the latent community structure.

\textbf{Notations.} Throughout the article, $X_n\pto X$ denotes convergence in probability, and $a_n\sim b_n$ indicates $\lim_{n\rightarrow\infty} a_n/b_n\rightarrow 1$.

\section{Sparse and scale-free networks}
\label{sec:sparsedef}

We first provide a formal definition of sparsity and scale-freeness, as there is no general agreement on the definition of a scale-free network and these notions are core to the results of this paper.

Let $(G_n)_{n\geq 1}$ be a sequence of simple random graphs of size $n$ where $G_n=(V_n,E_n)$, $V_n=\{1,\ldots,n\}$ is the set of vertices and $E_n$ the set of edges. Denote $|E_n|$ the number of edges. The graph is said to be sparse if $\mathbb E (|E_n|)/n^2\rightarrow 0$.  Let $N_k^{(n)}$ be the number of nodes of degree $k$ in $G_n$.
We now formally give the definition of a scale-free network informally introduced in Section~\ref{sec:introduction}.
\begin{defn}\label{def:scalefree}
A random graph sequence $(G_n)_{n\geq 1}$ is said to be scale-free with exponent $\eta$ iff there exists a slowly varying function $\ell$ and $\eta>1$ such that, for each $k=1,2,\ldots$
\begin{align}
\frac{N^{(n)}_k}{n}\pto \pi_k
\end{align}
as $n$ tends to infinity, where
\begin{equation}
\pi_k\sim\ell(k) k^{-\eta}\text{ as }k\rightarrow\infty.\label{eq:regvardegree}
\end{equation}
\end{defn}
Background definitions and properties of slowly and regularly varying functions are given in the appendix. Intuitively, slowly varying functions are functions that vary more slowly than any power of $x$. The term scale-free comes from the fact that the asymptotic degree distribution satisfies some (asymptotic) scale-invariance. For any integer $m\geq 1$,
\[
\lim_{k\rightarrow\infty}\frac{\pi_{mk}}{\pi_k}= m^{-\eta}.
\]
The most classical case is when $\ell(k)=C$ is constant. In this case, the asymptotic degree distribution behaves as a pure power-law for $k$ large. More generally, the scale-invariance property defined above will be satisfied for any slowly varying function $\ell$, which can be e.g. logarithm, or iterated logarithm. \cref{def:scalefree} is slightly more restrictive than the definition of a scale-free graph sequence in \citep[Definition 1.4]{vanderHofstad2016}, which is implied from \cref{def:scalefree} by properties of regularly varying functions (see appendix).

\section{Rank-1 inhomogeneous random graphs}
\label{sec:rank1}

\newcommand{\sn}{^{(n)}}

\subsection{Definition}
Let $(G_n)_{n\geq 1}$ be a sequence of simple random graphs of size $n$ defined as follows. The probability that two nodes $i$ and $j$ are connected in the graph $G_n$ is given by
\begin{equation}
p_{ij}^{(n)}=1-\exp\left ( -\frac{w_iw_j}{s^{(n)}}\right )\label{eq:norros-reittu}
\end{equation}
where $s\sn=\sum_{i=1}^n w_i$ and the positive weights $(w_1,w_2,\ldots)$ are independently and identically distributed (iid) from some distribution $F$ with $\mathbb E(w_1)<\infty$. The model~\eqref{eq:norros-reittu} is known as the Norros-Reittu (NR) inhomogeneous random graph model~\citep{Norros2006}. This model has been the subject of a lot of interest in the applied probability and graph theory literature~\citep{Bollobas2007,Bhamidi2012,vanderHofstad2013,vanderHofstad2016,Broutin2018}. The parameter $w_i>0$ accounts for degree heterogeneity in the graph and can be interpreted as a sociability parameter of node $i$. The larger this parameter, the more likely node $i$ is to connect to other nodes. 

\subsection{Sparsity and scale-free properties}

The random graph sequence defined by Equation \eqref{eq:norros-reittu} satisfies a number of remarkable asymptotic properties. The first result, which follows from \cite{Bollobas2007} (see details in the appendix), shows that the resulting graphs are sparse.

\begin{thm}[\cite{Bollobas2007}]\label{thm:rank1_sparsity}
Let $|E_n|$ denote the number of edges in the graph $G_n$. Then
\begin{align}
\frac{\mathbb E(|E_n|)}{n}\rightarrow  \frac{\mathbb E (w_1)}{2}~~\text{ and }~~
\frac{|E_n|}{n}\pto \frac{\mathbb E (w_1)}{2}.
\end{align}
\end{thm}

The following result is a corollary of Theorem 3.13, remark 2.4 and the discussion in Section 16.4 in \citep{Bollobas2007}. It states that the asymptotic degree distribution is a mixture of Poisson distributions, with mixing distribution $F$.
\begin{thm}[\cite{Bollobas2007}]\label{thm:rank1_degree}
Let $N^{(n)}_k$ be the number of vertices of degree $k$ in the graph $G_n$ of size $n$ and link probability $p_{ij}^{(n)}$ given by Equation~\eqref{eq:norros-reittu}. Then, for each $k=1,2,\ldots$, $N^{(n)}_k/n\pto \pi_k$ as $n$ tends to infinity, where
\begin{align}
 \pi_k:= \int_0^\infty \frac{x^k}{k!}e^{-x}dF(x).
\end{align}
\end{thm}

Our analysis on the asymptotic degree distribution is based on the following theorem
for the asymptotic behavior of mixed Poisson distributions.
\begin{thm}
\label{thm:mixed_poisson_asymp}
\citep{Willmot1990} Suppose that
\[
f(x) \sim \ell(x) x^\eta e^{-\zeta x}, \quad x \to \infty,
\]
where $\ell(x)$ is a locally bounded function on $(0,\infty)$ which varies slowy at infinity,
$\zeta \geq 0$, and $-\infty < \eta < \infty$ (with $\eta < -1$ when $\zeta=0$). For $\lambda > 0$, define the probabilities of the mixed Poisson distribution as
\[
\pi_k = \int_0^\infty \frac{(\lambda x)^k e^{-\lambda x}}{k!} f(x) dx; \quad k=0, 1, 2, \dots.
\]
Then,
\[
\pi_k \sim \frac{\ell(k)}{(\lambda + \zeta)^{\eta+1}} \bigg( \frac{\lambda}{\lambda+\zeta}\bigg)^k k^\eta, \quad k \to \infty.
\]
\end{thm}
The following result is a corollary of \cref{thm:rank1_degree} and \cref{thm:mixed_poisson_asymp}. It states that if the random variables $w_i$ are regularly varying (see definition in the appendix), then the sequence of random graphs is scale-free.

\begin{cor}\label{thm:scalefreemixed}
Let $N^{(n)}_k$ be the number of vertices of degree $k$ in the graph $G_n$ of size $n$ and link probability $p_{ij}^{(n)}$ given by Equation~\eqref{eq:norros-reittu}. Assume that the distribution F is absolutely continuous with pdf $f$ verifying $f(w)\sim \ell(w)w^{-\eta}$
as $w$ tends to infinity, for some locally bounded slowly varying function $\ell$ and $\eta> 1$.  Then, for each $k=1,2,\ldots$, $N^{(n)}_k/n\pto \pi_k$ as $n$ tends to infinity, where
\begin{align}
 \pi_k\sim \ell(k)  k^{-\eta}, \quad k \to \infty.
\end{align}
\end{cor}

\subsection{Particular examples}
We now consider two special cases. The first case yields scale-free graphs with asymptotic power-law degree distributions with exponent $\eta>2$. The second yields non-scale-free graphs, where the asymptotic degree distribution is power-law with exponential cut-off.
\subsubsection{Scale-free graph with power-law degree distribution}
\begin{figure*}
\centering
\includegraphics[width=0.48\linewidth]{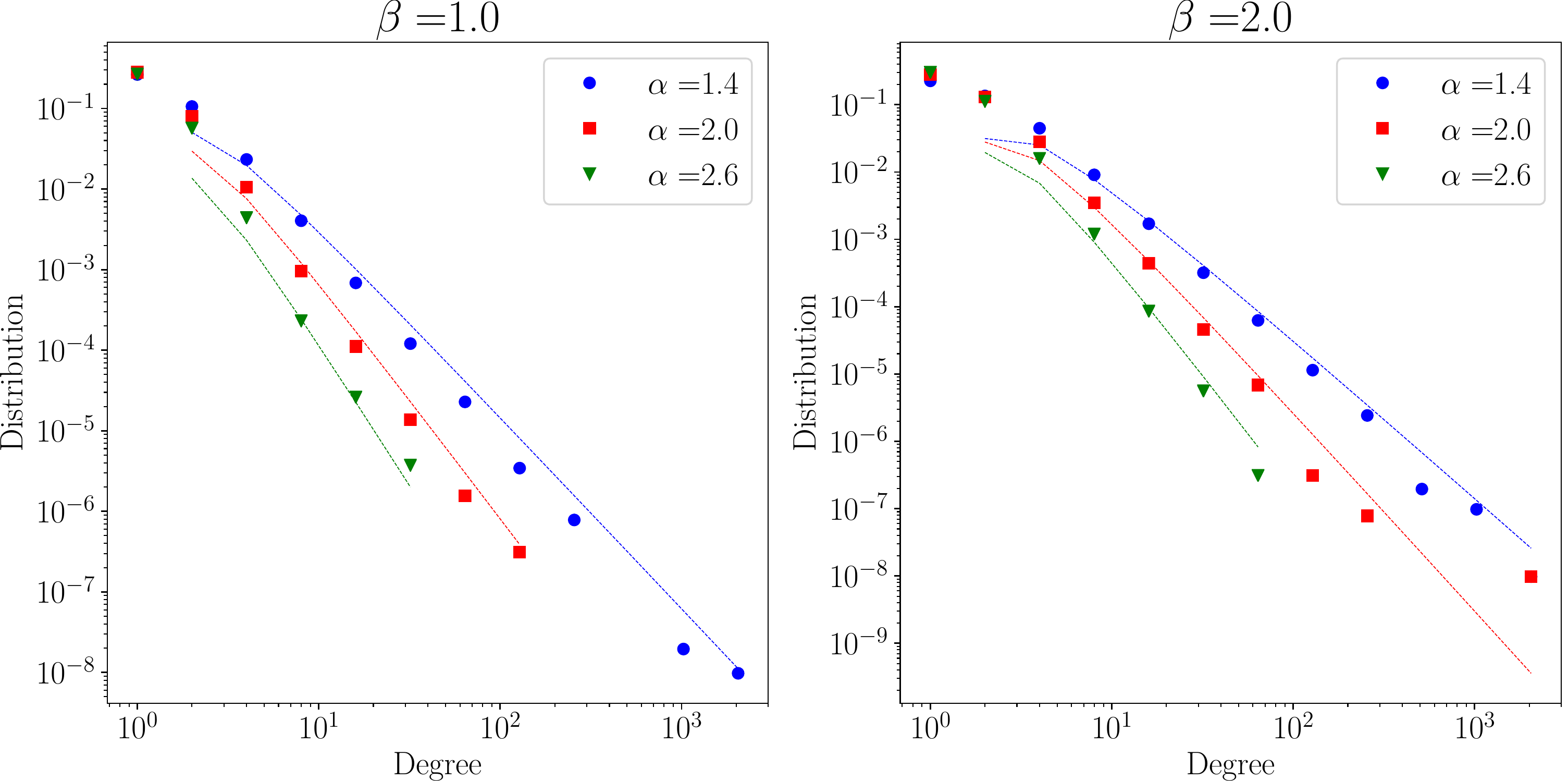}
\includegraphics[width=0.48\linewidth]{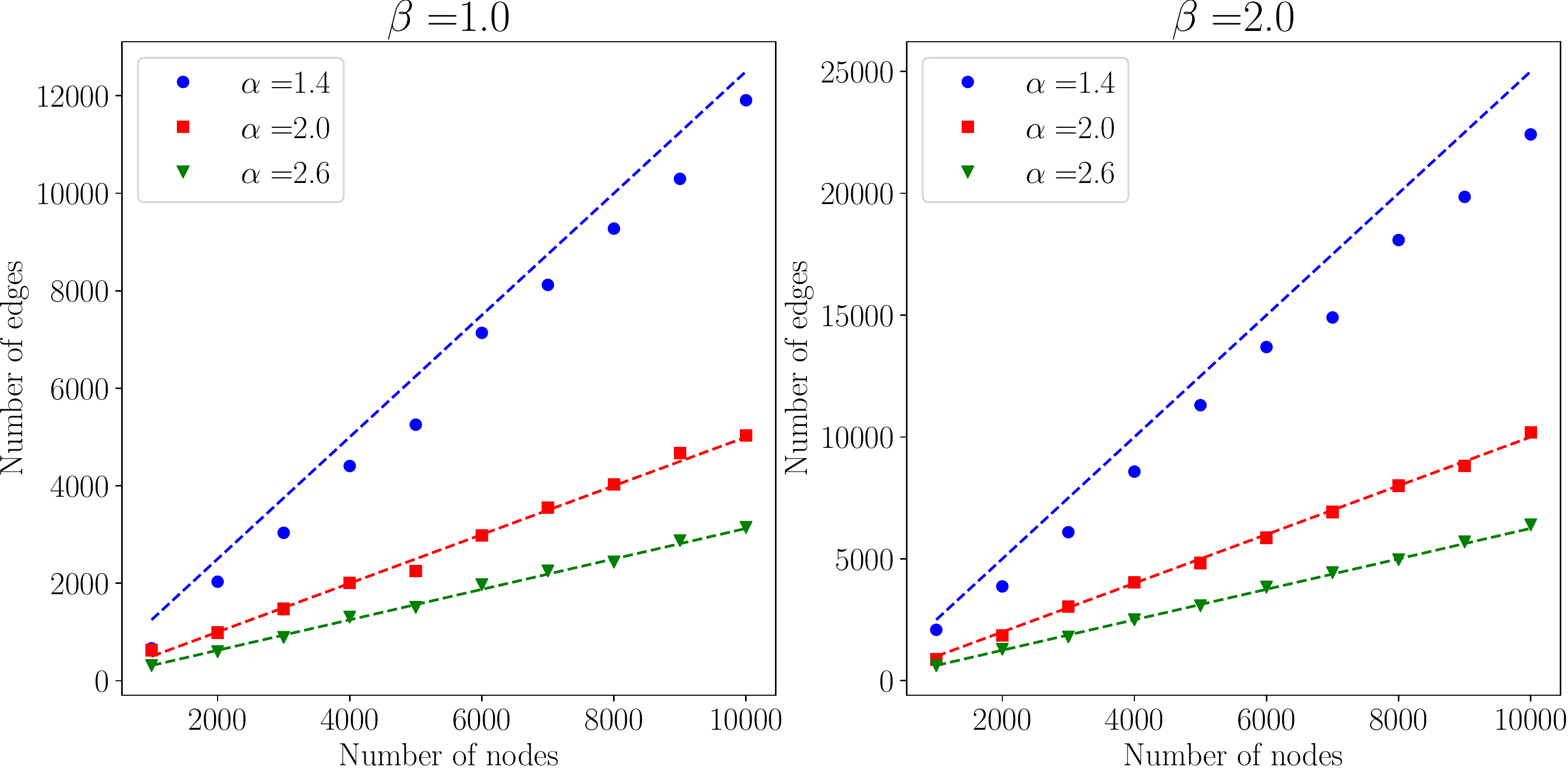}
\includegraphics[width=0.48\linewidth]{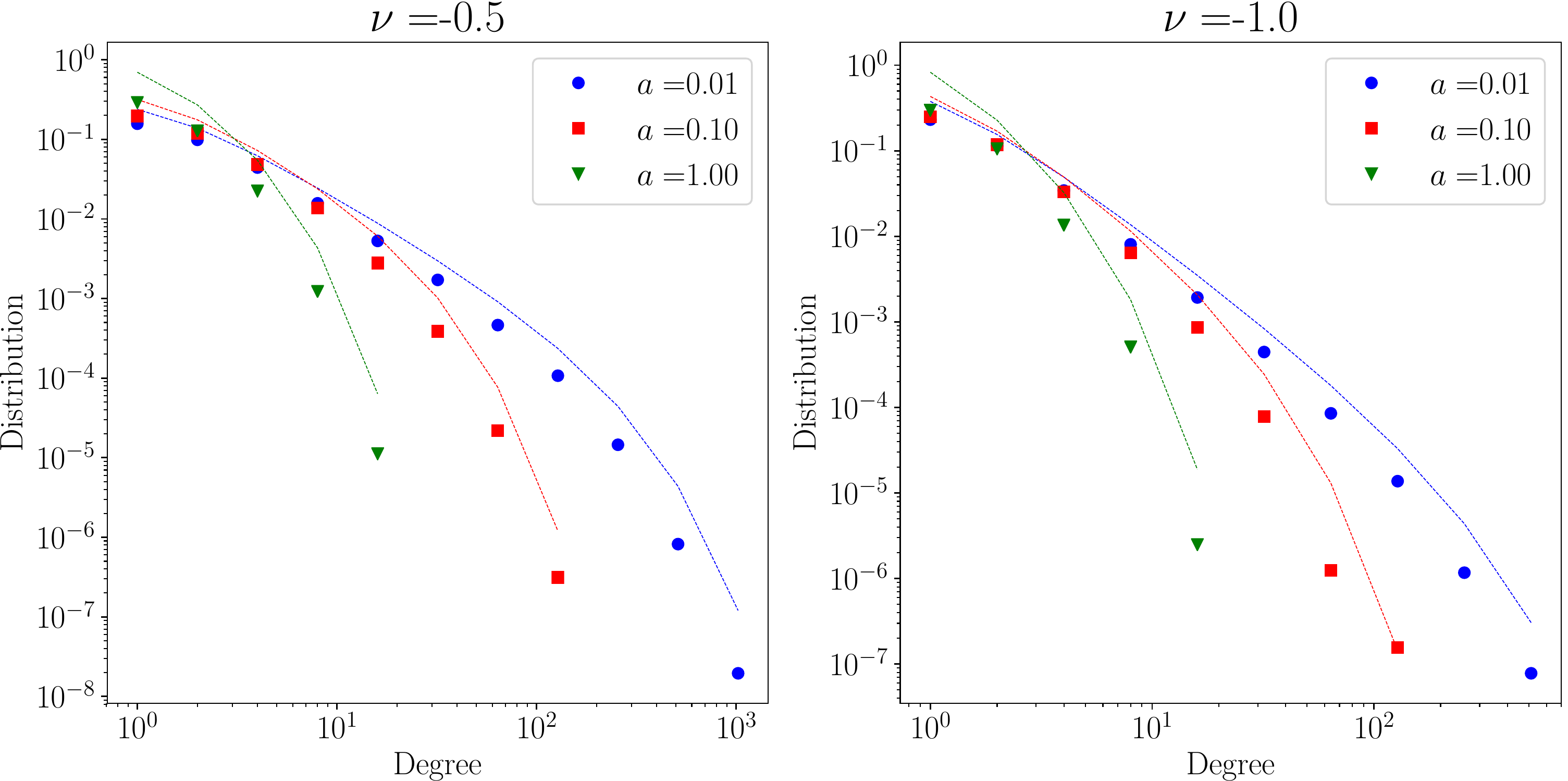}
\includegraphics[width=0.48\linewidth]{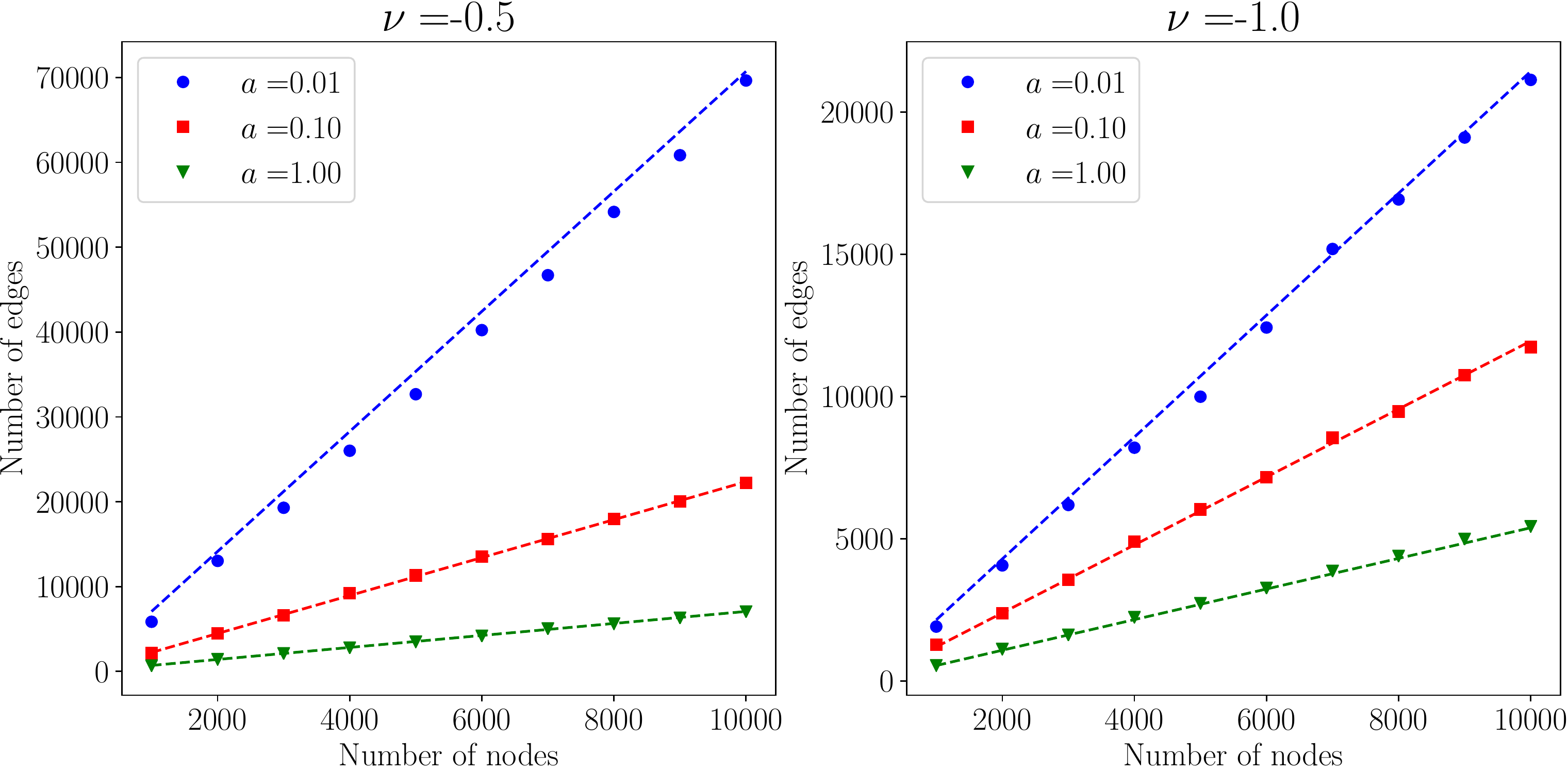}
\caption{
First row, first and second boxes: empirical degree distributions (dashed lines) of graphs with 10,000 nodes sampled from inverse gamma NR model  compared to the
theoretically expected asymptotic degree distribution (dotted lines), with various values of $\alpha$ and $\beta$. First row, third and fourth boxes: empirical number of edges (dashed lines) of graphs sampled from inverse gamma NR model versus the number of nodes compared to the
theoretically expected value of number of edges (dotted lines), with various values of $\alpha$ and $\beta$. Second row: the same figures for GIG NR model with various values of $\nu$ and $a$ with fixed $b=2.0$. Best viewed magnified in color.}
\label{fig:rank1_degree_sparsity}
\end{figure*}

For $i=1,2,\ldots$, let $w_i \iidsim \mathrm{invgamma}(\alpha, \beta)$ where $\mathrm{invgamma}(\alpha, \beta)$ denotes the inverse gamma distribution with parameters $\alpha>1$ and $\beta>0$, whose probability density function (pdf) is given by
$$
f(w)=\frac{\beta^\alpha}{\Gamma(\alpha)} w^{-\alpha-1} e^{-\beta/w}.
$$
Here, the constraint $\alpha > 1$ is required for the condition $\bbE[w_1]< \infty$. By
\cref{thm:rank1_degree}, the asymptotic degree distribution is a mixed Poisson-inverse-gamma distribution with probability mass function
\[
\pi_k = \frac{2\beta^{\frac{k+\alpha}{2}}}{k!\Gamma(\alpha)}K_{k-\alpha}(2\sqrt{\beta}) ,
\]
where $K$ is the modified Bessel function of the second kind. Using \cref{thm:scalefreemixed}, we obtain
\[
\pi_k \sim \frac{\beta^\alpha }{\Gamma(\alpha)} k^{-\alpha-1}
\quad \textrm{ as }k \to \infty.
\]
The resulting asymptotic degree distribution is a power-law and the graph is scale-free with arbitrary index $\alpha + 1 > 2$. The two hyperparameters of the inverse gamma prior play an important role to decide the asymptotic properties of graphs. The shape parameter $\alpha$ tunes the index of power-law, and is also related to the sparsity of graphs. The scale parameter $\beta$ is also related to the sparsity of graphs. \cref{fig:rank1_degree_sparsity} shows the empirical degree distributions and number of edges of graphs generated from inverse gamma NR model.

\subsubsection{Non scale-free graph with power-law degree distribution with exponential cut-off}
Now we consider another model with generalized inverse Gaussian (GIG) prior. Let $w_i \iidsim \mathrm{GIG}(\nu, a, b)$ where the density of the GIG distribution with parameter $\nu$, $a>0$ and $b>0$ is given by
\[
f(w) = \frac{(a/b)^{\nu/2}}{2K_\nu(\sqrt{ab})} w^{\nu-1} \exp\bigg\{
-\frac{1}{2}\bigg(aw + \frac{b}{w}\bigg)\bigg\}.
\]
Note that by taking $a\rightarrow 0$, one obtains the pdf of an inverse gamma distribution as a limiting case.
By \cref{thm:rank1_degree}, the asymptotic degree distribution is
\[
\pi_k = \frac{(a/b)^{\nu/2}}{k!\{(a+2)/b\}^{(k+\nu)/2}}
\frac{K_{k+\nu}(\sqrt{(a+2)b})}{K_\nu(\sqrt{ab})}.
\]
This distribution is sometimes called the Sichel distribution, after Herbert Sichel~\citep{Sichel1974}. Note that $f(w)\sim (a/b)^{\nu/2}/2/K_\nu(\sqrt{ab}) w^{\nu-1} \exp(-aw/2)$ as $w\rightarrow\infty$ hence, by \cref{thm:mixed_poisson_asymp},
\[
\pi_k \sim \frac{(a/b)^{\nu/2} k^{\nu-1} e^{-\log(1+a/2)k} }{2(1 + a/2)^\nu K_\nu(\sqrt{ab})}
 \quad \textrm{ as } k \to \infty.
\]
In this case, the asymptotic degree distribution is not of the form of Equation \eqref{eq:regvardegree}, and the graph sequence is therefore not scale-free. However, the asymptotic degree distribution has the form $k^{\nu-1}e^{-\tau k}$ of a power-law distribution with exponential cut-off. This class of probability distributions has been shown to provide a good fit to the degree distributions of a wide range of real-world networks~\citep{Clauset2009}.
As for the inverse gamma NR model, the hyperparameters $(\nu, a, b)$ tunes the asymptotic properties. $\nu$ determines the power-law index of degree distribution, $a$ is related
to the exponential cutoff and sparsity, and $b$ is related to the sparsity. \cref{fig:rank1_degree_sparsity} shows the empirical degree distributions and the number of edges of graphs generated from GIG NR model.

\section{Extension to Latent Overlapping Communities}
\label{sec:rankc}
\subsection{Definition}
The inhomogeneous random graphs considered so far only account for degree heterogeneity. However, the connections in real-world networks are often due to some latent interactions between the vertices. Recently, several models that combine a degree correction together with a latent structure to define edge probabilities were proposed~\citep{Zhou2015,Todeschini2016, Herlau2016, Lee2017}.
In this section, we propose an extension of the NR model that includes some latent overlapping structure, and study the sparsity, scale-freeness properties and asymptotic degree distribution of this model. Let the edge probability between the vertex $i$ and $j$ be given by
\[\label{eq:rankc_link}
p^{(n)}_{ij} = 1 - \exp\bigg( - \frac{w_iw_j}{s\sn }\sum_{q=1}^c
\frac{v_{iq} v_{jq}}{r_q\sn/n}\bigg).
\]
where $(w_i)_{i=1,2,\ldots}$ are iid random variables with distribution $F$ with $\mathbb E(w_1)<\infty$ and $(v_{i1},\ldots,v_{ic})_{i=1,2,\ldots}$ are i.i.d. with $\mathbb E(v_{1q})<\infty$ for all $q$ and
$r_q\sn = \sum_{i=1}^n v_{iq}$. We call this model with $c$ communities the \emph{rank-$c$ model}. As in the rank-1 model, the parameter $w_i$ can be interpreted as an overall sociability parameter of node $i$, or degree-correction. The parameter $v_{iq}$ can be interpreted as the level of affiliation of individual of $i$ to community $q$. Similar models, in a different asymptotic framework have been used in ~\citep{Yang2013,Zhou2015,Todeschini2016}.


\begin{thm}\label{thm:rankc_sparsity_and_degree}
Let $|E_n|$ denote the number of edges in the graph $G_n$ defined with link probability \eqref{eq:rankc_link}. Then,
\[
\frac{\bbE(|E_n|)}{n} &\rightarrow \frac{\bbE(w_1) \sum_{q=1}^c \bbE(v_{1q})}{2}\\
\frac{|E_n|}{n} &\pto \frac{\bbE (w_1) \sum_{q=1}^c\bbE (v_{1q})}{2}.
\]
Recall that $N^{(n)}_k$ is the number of vertices of degree $k$ in the graph $G_n$ of size $n$. Then, for each $k=1,2,\ldots$, $N^{(n)}_k/n \pto \pi_k$ as $n$ tends to infinity, where
\begin{align}
 \pi_k= \int_0^\infty\int_0^\infty \frac{(uw)^k}{k!}e^{-uw}dF(w)dH(u)
\end{align}
where $H$ is the distribution of the random variable $U=\sum_{q=1}^c v_{1q}$. If additionally $F$ is absolutely continuous with pdf $f$ verifying
$
f(w)\sim \ell(w)w^{-\eta}$ as $w\rightarrow\infty$ for some locally bounded slowly varying function $\ell$ and $\eta> 1$ and $\mathbb E(U^{\eta-1+\epsilon})<\infty$ for some $\epsilon>0$, then
$$
\pi_k\sim \mathbb E(U^\eta)\ell(k)k^{-\eta}~~\text{as }k\rightarrow\infty.
$$
\end{thm}
The proof of \cref{thm:rankc_sparsity_and_degree} is given in the appendix.
In this paper, we consider in particular
\[
(v_{i1}, \dots, v_{iq}) \sim \mathrm{Dir}(\gamma),
\]
where $\mathrm{Dir}(\gamma)$ denotes the standard Dirichlet distribution with parameter $\gamma=(\gamma_1,\ldots,\gamma_c)$, where $\gamma_q>0$ for $q=1,\ldots,c$.


\section{Posterior inference}
\label{sec:inference}
\subsection{Posterior inference for the rank-1 NR}
Let $Y = \{y_{ij}\}_{1\leq i < j \leq n}$ be an (upper triangular part of) adjacency matrix of a graph $G_n$ and $w=(w_1,\ldots,w_n)$. The joint density is written as
\[
p(Y, w) = \prod_{i=1}^n f(w_i) \prod_{i<j} \Big(1-e^{-\frac{w_iw_j}{s\sn}}\Big)^{y_{ij}} e^{(y_{ij}-1)\frac{w_iw_j}{s\sn}}
\]
Following \cite{Caron2017} and \cite{Zhou2015}, we introduce a set of auxiliary truncated Poisson random variables $m_{ij}$ for the pairs with $y_{ij}=1$.
\[
p(m_{ij}|w) = \frac{(\frac{w_iw_j}{s\sn})^{m_{ij}}\exp(-\frac{w_iw_j}{s\sn})\indicator{m_{ij}>0}
}{m_{ij}!(1-\exp(-\frac{w_iw_j}{s\sn})}.
\]
The log joint density is then given as
\[\label{eq:rank1_joint}
\log p(Y, M, w) = \sum_{(i,j)\in E_n} \bigg(m_{ij} \log \frac{w_iw_j}{s\sn} - \log m_{ij}!\bigg) 
 + \frac{1}{2} \bigg( \sum_{i=1}^n \frac{w_i^2}{s\sn} - s\sn\bigg) + \sum_{i=1}^n \log f(w_i).
\]
Note that the terms for the pairs without edges $(y_{ij}=0)$ are collapsed into a single summation, and hence the overall computations of the log joint density and its gradient take
$O(n + |E_n|)$ time. This is a huge advantage of the link function of NR model, while other link functions for rank-1 inhomogeneous random graphs~\citep{Britton2006,Chung2002, Chung2002a, Chung2003} suffer from $O(n^2)$ computing times.

For the posterior inference, we use a Markov chain Monte Carlo (MCMC) algorithm. At each step, given the gradient of the log joint density, we update $w$ via Hamiltonian Monte Carlo (HMC,~\citep{Duane1987,Neal2011}). Then we resample the auxiliary variables $m$ from truncated Poisson, and update hyperparameters for $f(w)$ using a Metropolis-Hastings step. Details can be found in the appendix.

\subsection{Posterior inference for the rank-$c$ NR}
The posterior inference for the rank-$c$ model is similar to that of the rank-1 model. Following \cite{Todeschini2016}, for tractable inference, we introduce a set of multivariate truncated Poisson random variables $M = ((m_{ijq})_{q=1}^c)_{(i,j)\in E_n}$,
\[
p(M|w, V) = \prod_{(i,j)\in E_n} \prod_{q=1}^c \frac{\lambda_{ijq}^{m_{ijq}} e^{-\lambda_{ijq}}
\indicator{\sum_{q'=1}^c m_{ijq'}>0}}{1 - \exp(-\sum_{q'=1}^c \lambda_{ijq'})}.
\]
where $\lambda_{ijq} = \frac{w_iw_j}{s\sn} \frac{v_{iq}v_{jq}}{r_q\sn/n}$ and $V=(v_{iq})_{i=1,\ldots,n,q=1,\ldots,c}$. The log joint density is
\[
\log p(Y, M, w, V) &= \prod_{(i,j)\in E_n} \sum_{q=1}^c (m_{ijq} \log \lambda_{ijq} - \log m_{ijq}!) 
 - \sum_{i<j} \sum_{q=1}^c \lambda_{ijq} + \sum_{i=1}^n \log f(w_i) \nonumber\\
& + \sum_{i=1}^n \log g(v_{i1}, \dots, v_{ic};\gamma),
\]
where $g(\cdot;\gamma)$ is the density for Dirichlet distribution with parameters $\gamma$. As for the rank-1 model, we can efficiently compute this log joint density and its gradient w.r.t. $w$ and $V$ with $O(cn+c|E_n|)$ time. At each step of MCMC, we first sample $w$ and $V$ via HMC, resample $M$ from multivariate truncated Poisson, and update hyperparameters via Metropolis-Hastings. The detailed procedure can be found in the appendix.

\section{Discussion}
\label{sec:discussion}

The models described in this paper can capture sparsity, scale-freeness with exponent $\eta>2$ and latent community structure. One drawback of the construction is that the model lacks projectivity, due to normalisation by $s_n$ in the link probability~\eqref{eq:norros-reittu}. While this is an undesirable feature of the approach, we stress that there does not exist any projective class of random graphs that can capture all those properties, as we explain below. A popular class of models is the graphon-based or vertex-exchangeable graphs, which include as special cases stochastic blockmodels, latent factor models and their extensions, see \citep{Orbanz2015} for a review. While these models have been successfully applied in a wide range of application, they produce dense graphs with probability one, as stressed by \cite{Orbanz2015}. Alternative models have been proposed, either based on exchangeable point processes~\citep{Caron2017,Veitch2015,Borgs2016}, or on the notion of edge-exchangeability~\citep{Crane2015,Crane2017,Cai2016}. \citet{Caron2017a} showed that using exchangeable point processes, one can obtain scale-free graphs with exponent $\eta\in(1,2]$, but not above. While no results exist for the scale-freeness of edge-exchangeable random graphs in the sense of \cref{def:scalefree} (see \cite[Problem 9.8]{Janson2017}), it is likely that a similar range is achieved for this class of models. Another family of models are non-exchangeable models based on preferential attachment~\citep{Barabasi1999}. The generated graphs are scale-free with exponent $\eta>2$. However, the generative process makes it difficult to consider more general constructions that take into account community structure. Additionally, the non-exchangeability implies that the ordering of nodes must be known or need to be inferred for inference, which limits its applicability. By contrast, our model is finitely exchangeable for each $n$, and so the ordering of the nodes needs not to be known in order to make inference. As a consequence, no other projective class of model can give scale-free networks with exponent $\eta>2$, interpretable parameters capturing community structure, and scalable inference, as described in this paper. While the model has a number of attractive properties, it also has some limitations. The mean number of triangles in inhomogeneous random graphs converges to a constant as $n$ tends to infinity~\citep{vanderHofstad2018}. Although the latent community structure introduced may mitigate this effect for reasonable $n$, this property appears undesirable for real-world network.

\section{Experiments}
\label{sec:experiments}
\subsection{Experiments with the rank-1 models}
In this section, we test our inverse-gamma NR model (IG-NR) and generalized inverse Gaussian NR model (GIG-NR) on synthetic and real world graphs. For all experiments, we ran three MCMC chains for 10,000 iterations for our algorithms, and collected every 10th samples after 5,000 burn-in samples. The prior distributions for the hyperparameters of the different models are given in the appendix. The code for our experiments is available at \url{https://github.com/OxCSML-BayesNP/BNRG}.

\begin{figure*}
\centering
\includegraphics[width=0.24\linewidth]{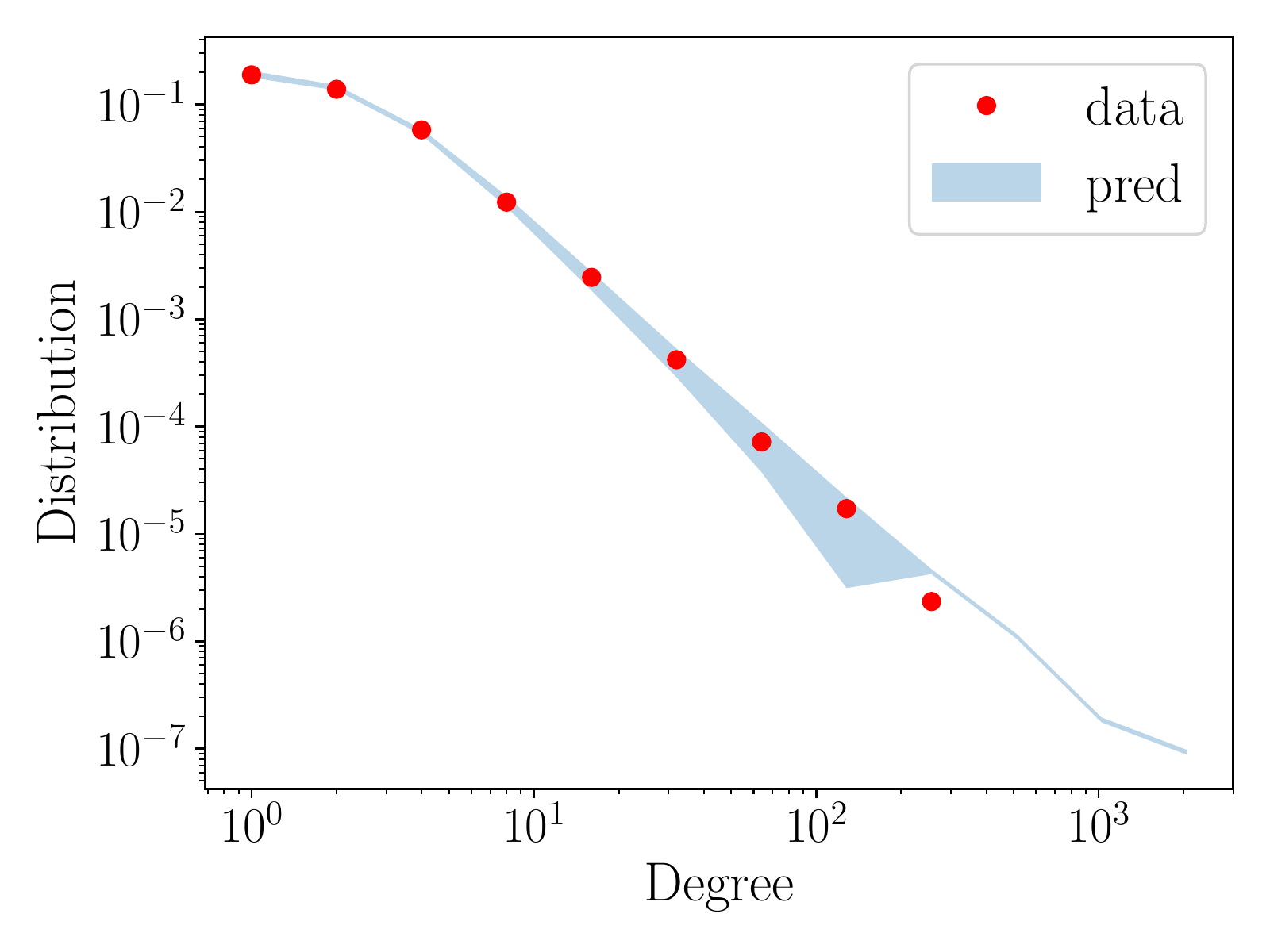}
\includegraphics[width=0.24\linewidth]{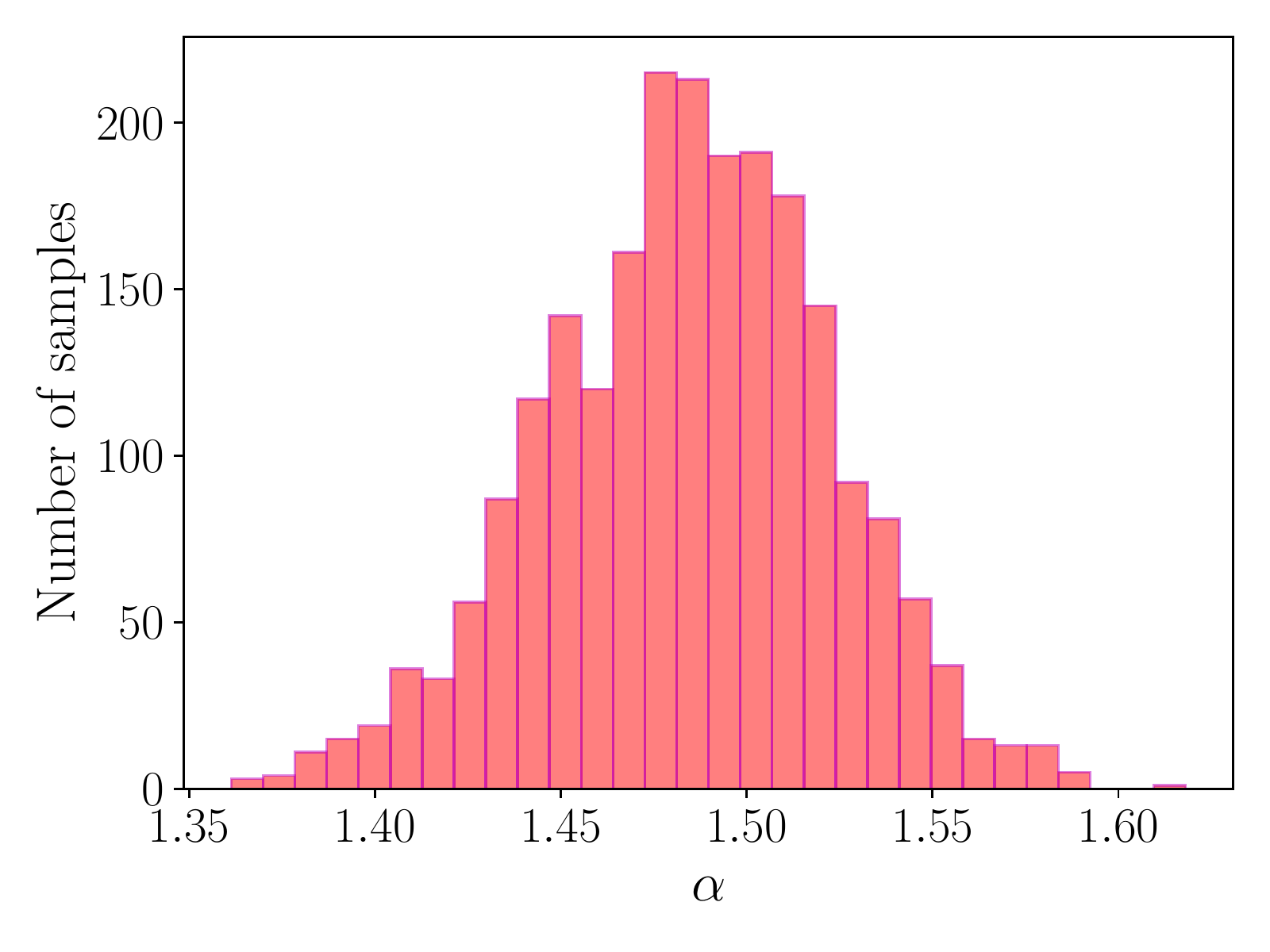}
\includegraphics[width=0.24\linewidth]{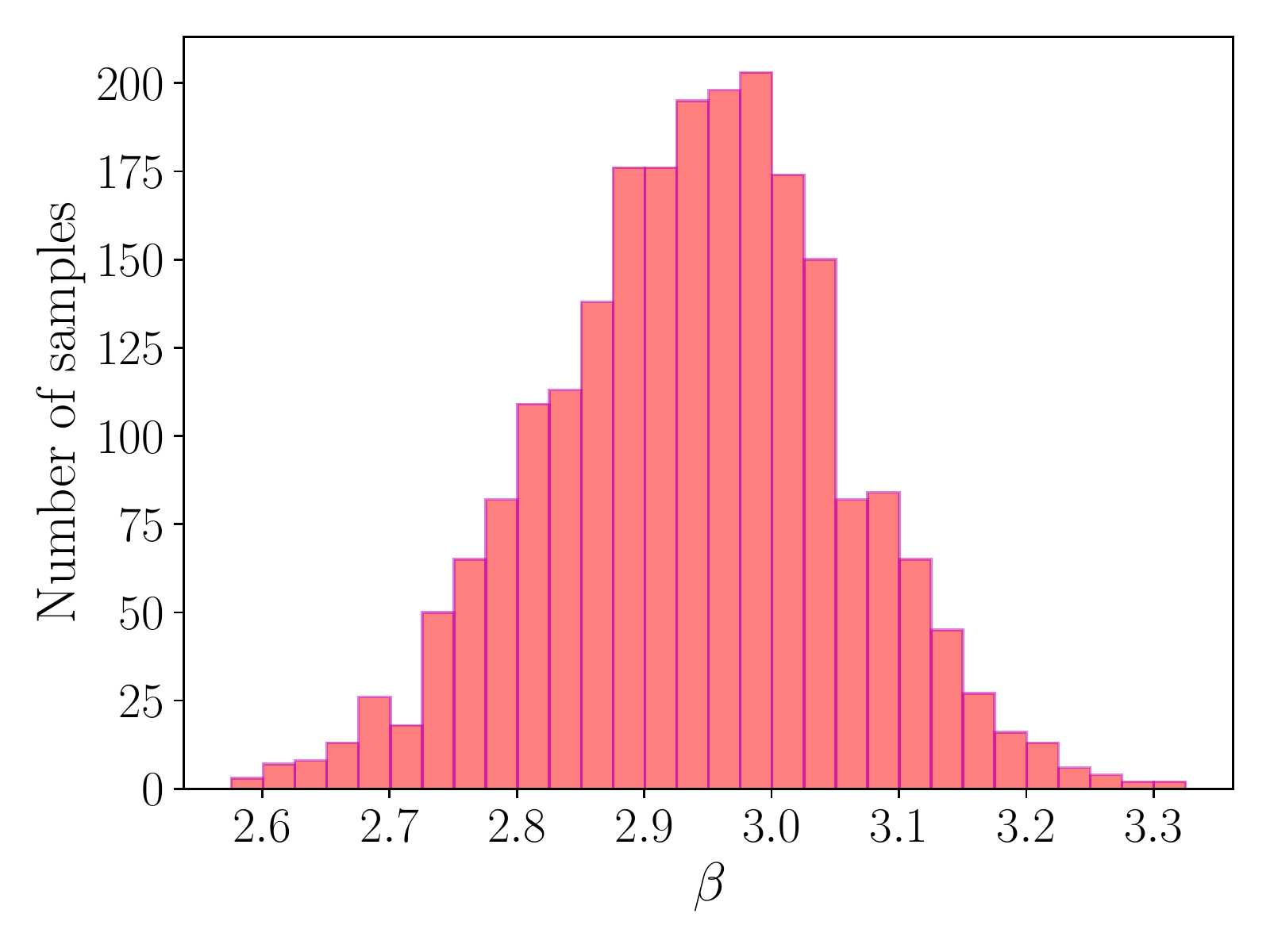}\newline
\includegraphics[width=0.24\linewidth]{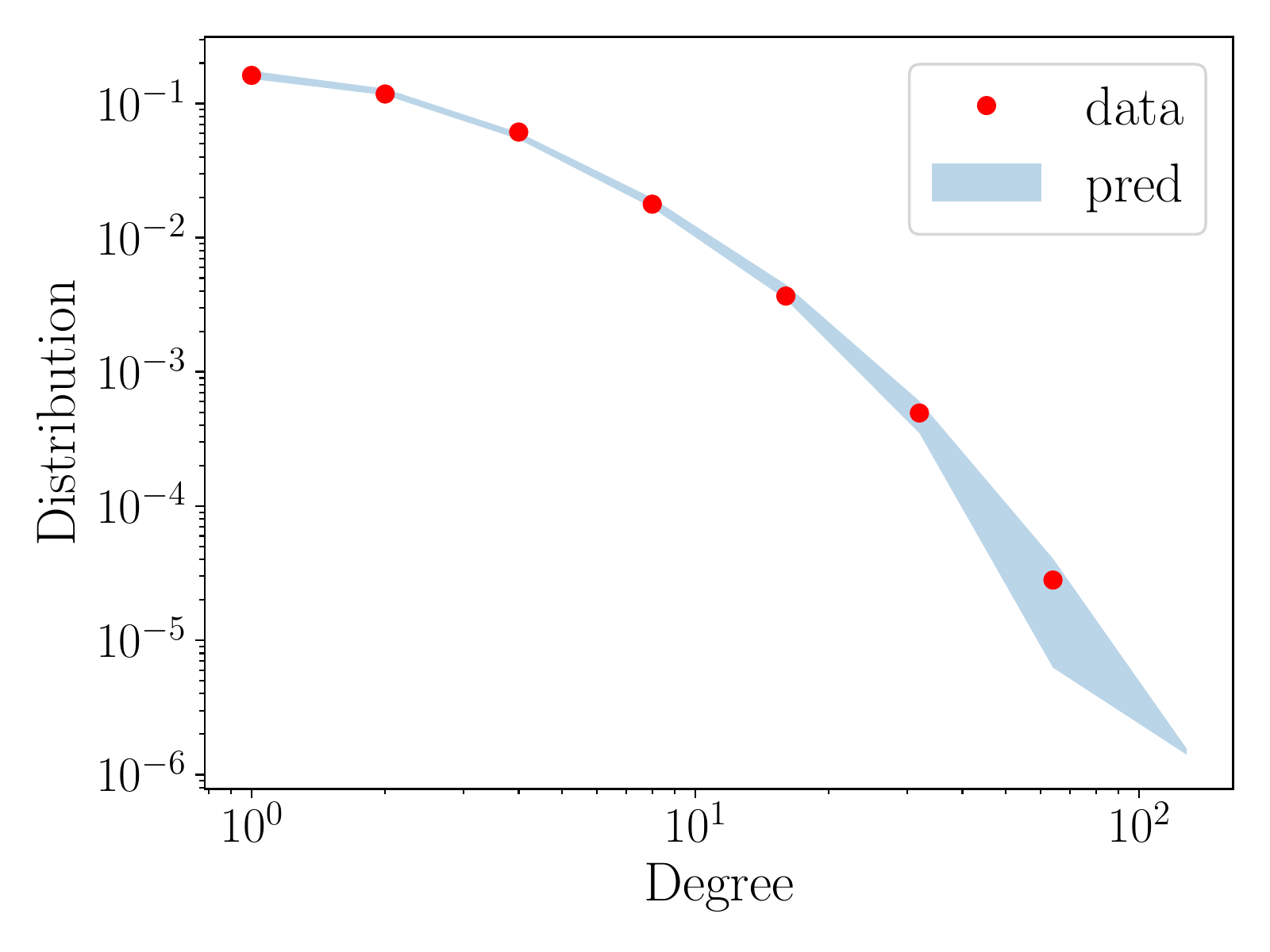}
\includegraphics[width=0.24\linewidth]{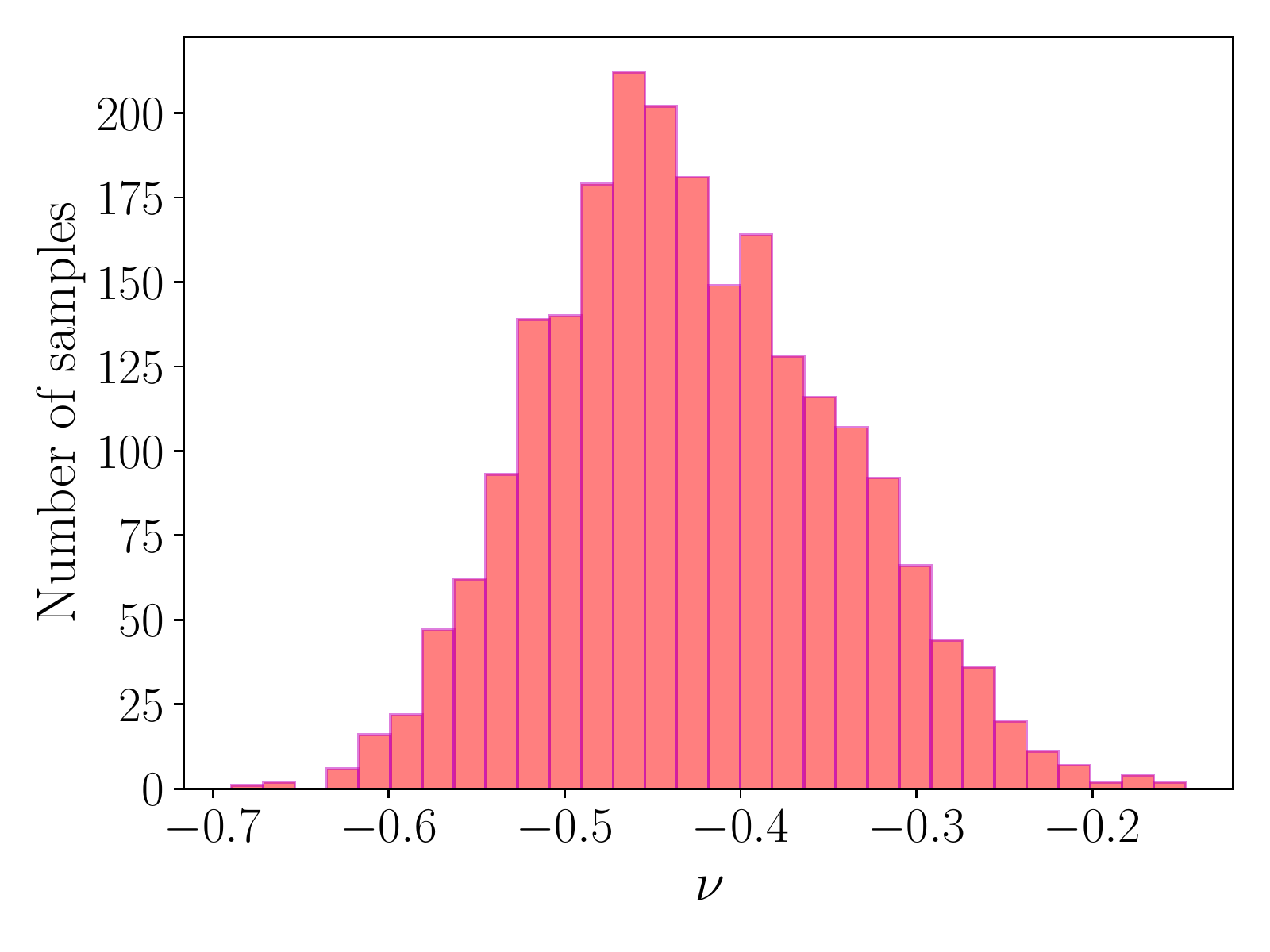}
\includegraphics[width=0.24\linewidth]{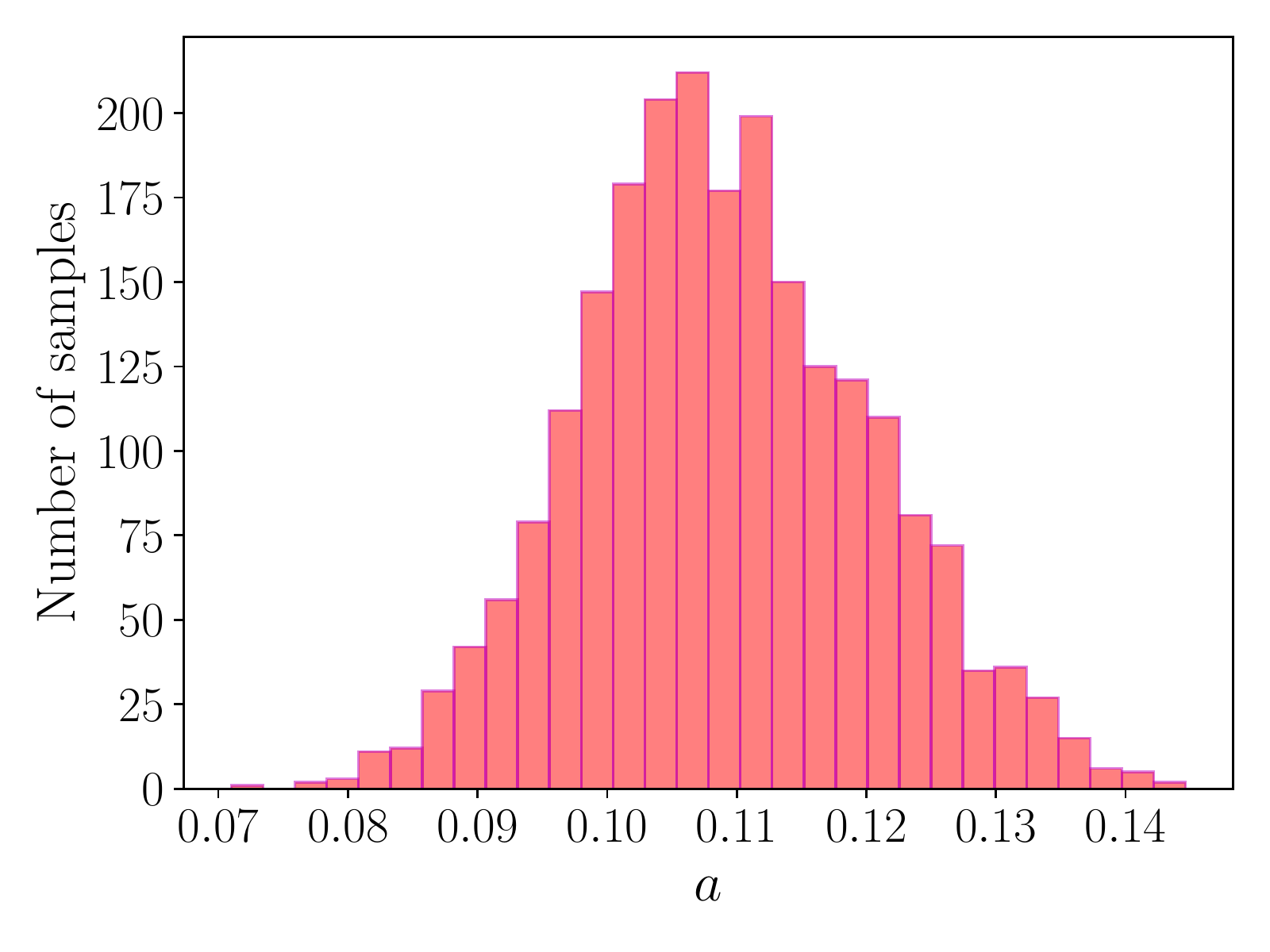}
\includegraphics[width=0.24\linewidth]{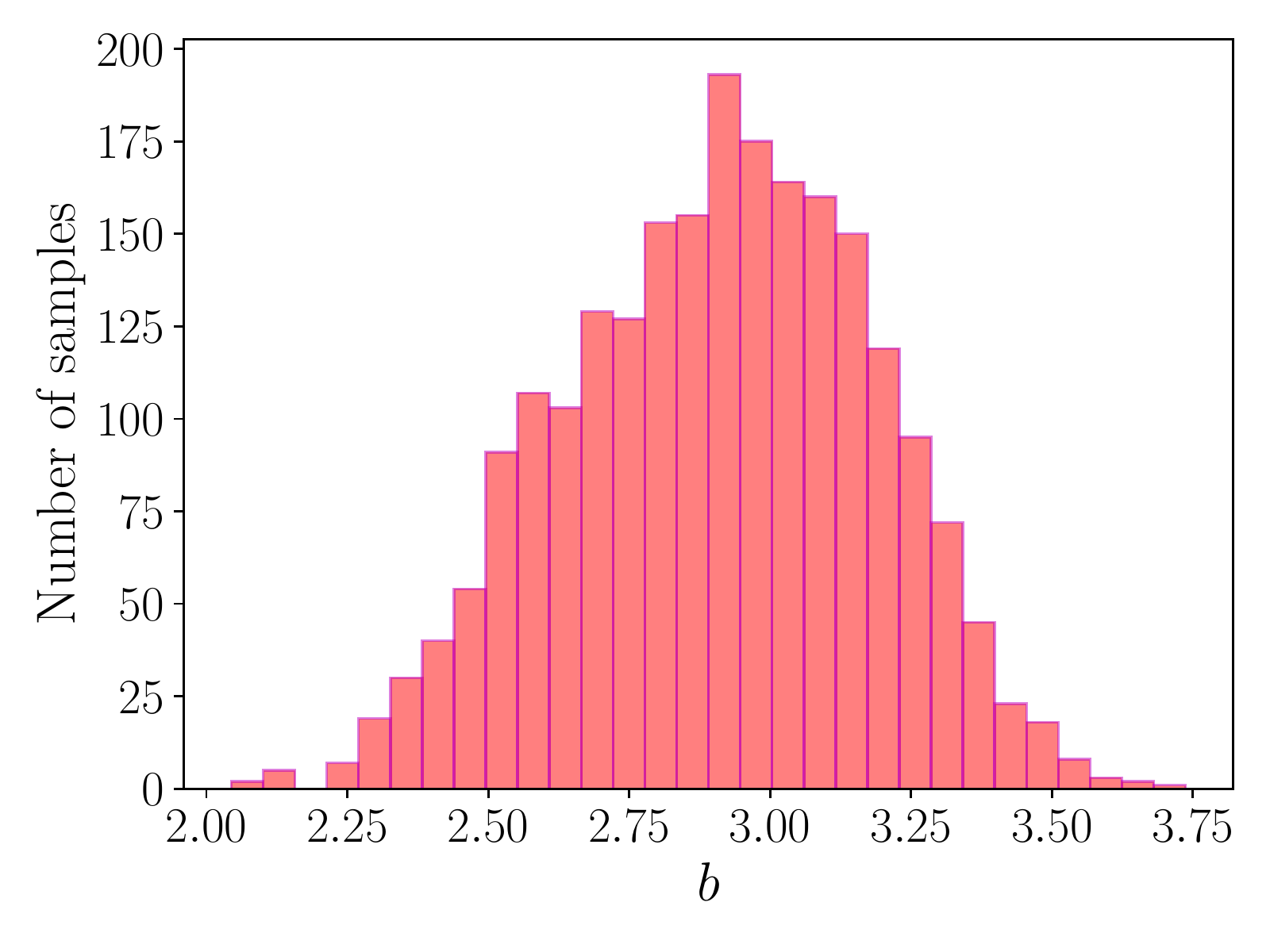}
\caption{(Top row): IG-NR. 95$\%$ credible intervals of the predictive
degree distribution, posterior samples of the hyperparameters $\alpha$ and $\beta$. The true values are $\alpha=1.5$ and $\beta=3.0$. (Bottom row): GIG-NR. 95$\%$ credible intervals of the predictive
degree distribution and posterior samples of the hyperparameters $\nu$, $a$ and $b$. The true values are $\nu=-0.5, a=0.1$ and $b=3.0$.}
\label{fig:synth}
\end{figure*}

\paragraph{Experiments with synthetic graphs.} We first fitted the basic models with Inverse-gamma prior (IG) and generalized inverse Gaussian prior (GIG) on synthetic graphs generated from IG-NR model and GIG-NR model. For IG, we generated a graph with $n=5,000$ nodes with parameters $\alpha=1.5$ and $\beta=3.0$. For GIG, we
generated a graph with 5,000 nodes with parameters $\nu=0.5, a=0.1, b=3.0$. As summarized in
\cref{fig:synth}, the posterior distribution recovers the hyperparameter values used to generated the graphs, and the posterior predictive distribution provides a good fit to the empirical degree distribution.

\begin{figure}
\centering
\subfigure[\texttt{cond-mat}]{\includegraphics[width=.32\linewidth]{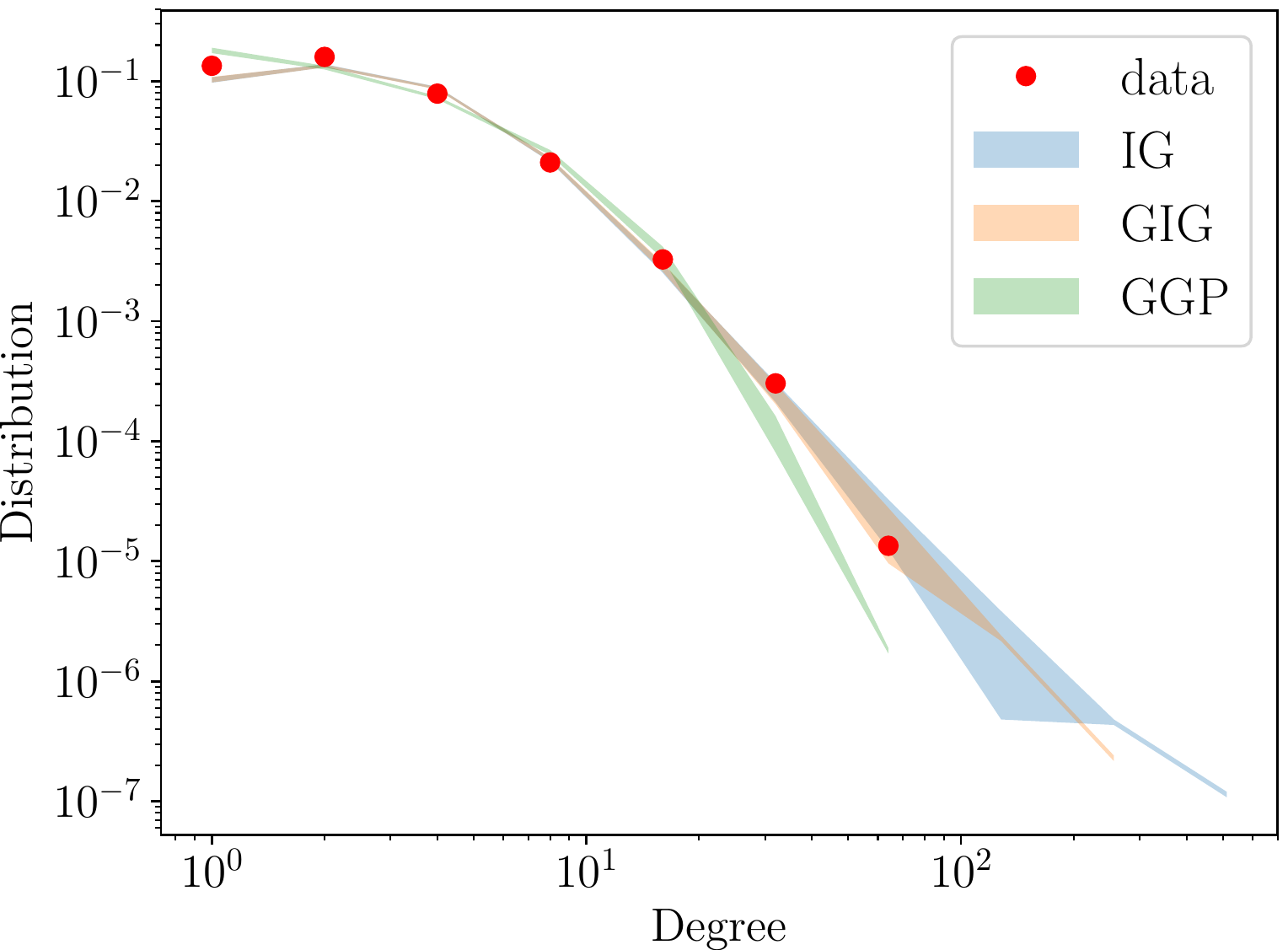}}
\subfigure[\texttt{Enron}]{\includegraphics[width=.32\linewidth]{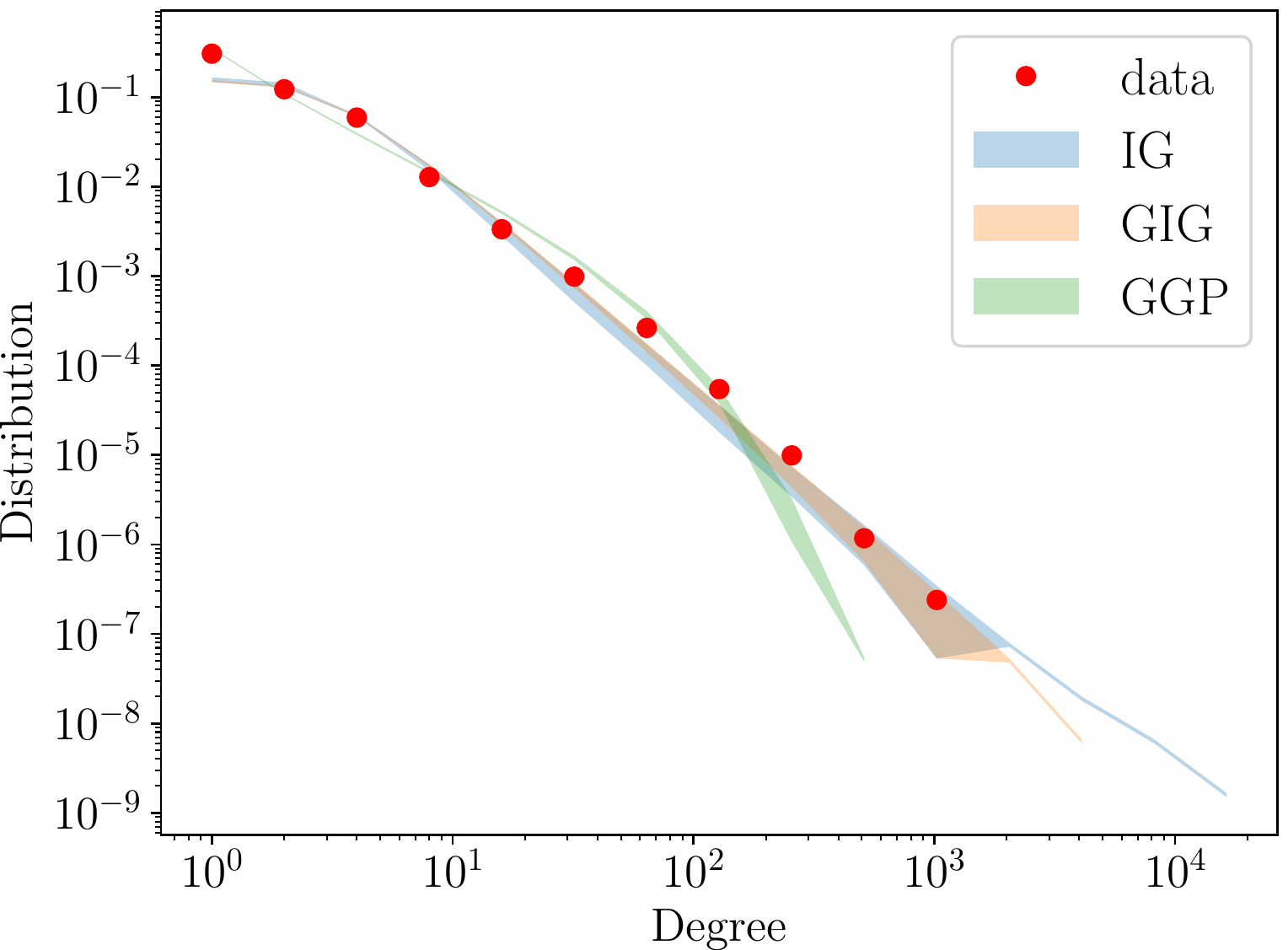}}
\subfigure[\texttt{internet}]{\includegraphics[width=.32\linewidth]{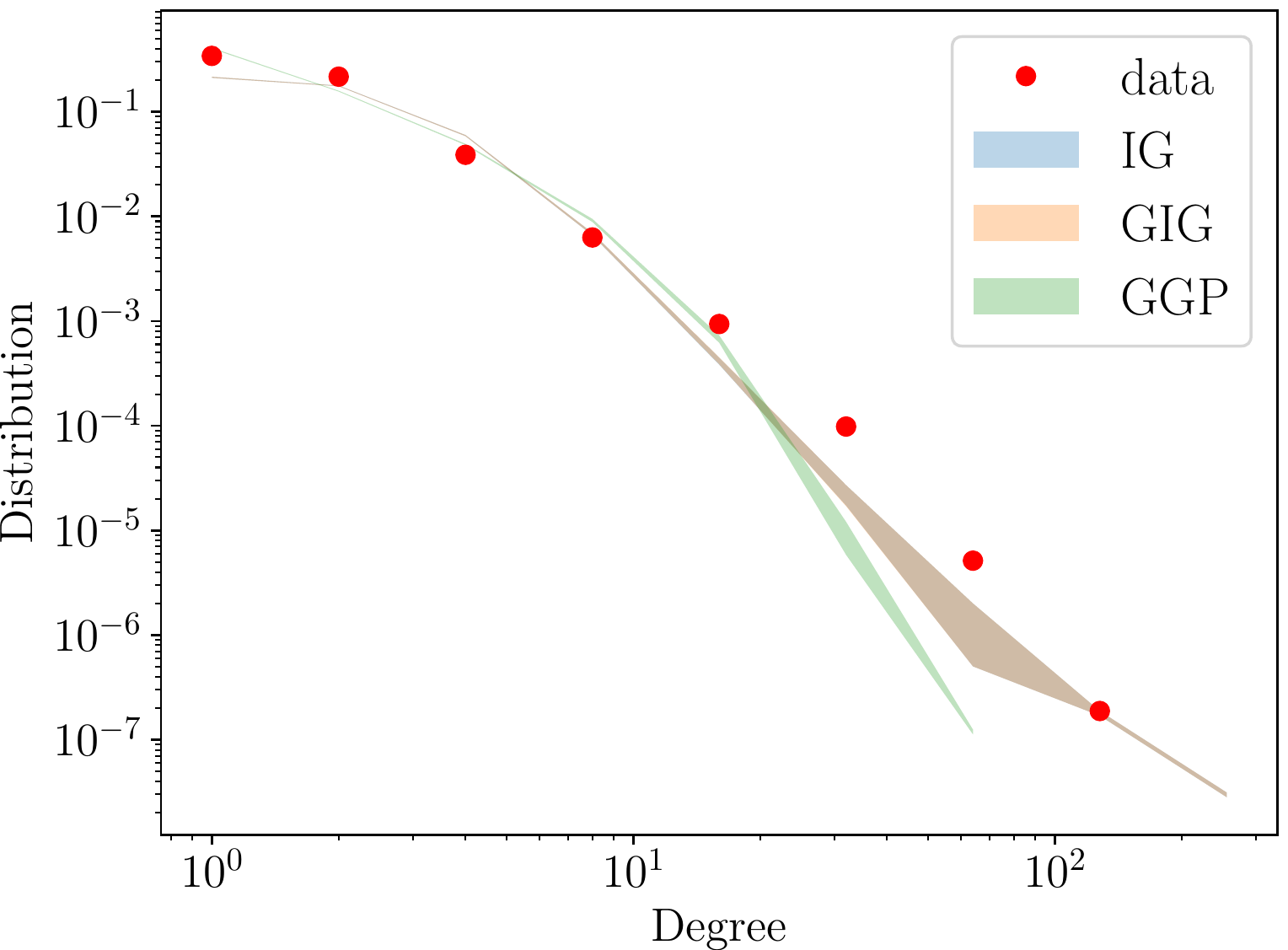}}
\caption{95\% credible intervals of predictive degree distributions of IG, GIG, GGP prior random graph models on (a) \texttt{cond-mat}, (b) \texttt{Enron} and (c) \texttt{internet} (right) graphs.}
\label{fig:rank1_real}
\end{figure}

\paragraph{Experiments with real-world graphs.} Now we evaluate our models on three real-world networks: \\
~~~$\bullet$ \texttt{cond-mat}\footnote{\url{https://toreopsahl.com/datasets/\#newman2001}}: co-authorship network based on arXiv preprints for condensed matter, 16,264 nodes and 47,594 edges.\\
$\bullet$ \texttt{Enron}\footnote{\url{https://snap.stanford.edu/data/email-Enron.html}}: Enron collaboration e-mail network,
36,692 nodes and 183,831 edges.\\
$\bullet$ \texttt{internet}\footnote{\url{https://www.cise.ufl.edu/research/sparse/matrices/Pajek/internet.html}}: Network of internet routers, 124,651 nodes and 193,620 edges.\\
To evaluate the goodness-of-fit in terms of degree distributions, as suggested in \citet{Clauset2009}, we sample graphs from the posterior predictive distribution based on the posterior samples, and computed the reweighted Kolmogorov-Sminorov (KS) statistic:
\[
D = \max_{x\geq x_{\min}} \frac{|S(x) - P(x)|}{\sqrt{P(x)(1-P(x))}},
\]
where $S(x)$ is the CDF of observed degrees, $P(x)$ is the CDF of degrees of graphs sampled from the predictive distribution, and $x_{\min}$ is the minimum $x$ values among the observed degree and predictive degree. We compare our model to the random graph model with generalized gamma process prior (GGP, \citep{Caron2017}), whose asymptotic degree distribution is a power-law with exponent in $(1, 2)$. We ran MCMC for the GGP model with 40,000 iterations and three chains. Posterior predictive degree distribution are reported in \cref{fig:rank1_real}. Credible intervals of the hyperparameters and KS statistics for the different models are given in \cref{tab:rank1_real}. Both IG and GIG provide a good to the degree distribution, with an exponent greater than 2, while the GGP model fails to capture the shape of 
the degree distribution.

\begin{table}\centering
\small
\caption{Average reweighted KS statistic of predictive degree distributions and 95\% credible intervals of estimated hyperparameters for IG, GIG and GGP models.}
\begin{tabular}{@{}  c  c M{2.5cm} c M{2.5cm} c M{2.5cm}  @{}}\toprule
& \multicolumn{2}{c}{\texttt{cond-mat}} & \multicolumn{2}{c}{\texttt{Enron}} & \multicolumn{2}{c}{\texttt{internet}}\\
\cmidrule{2-3} \cmidrule{4-5} \cmidrule{6-7}
&$D$ & hyperparams &$D$ & hyperparams &$D$ & hyperparams \\ \midrule
IG  &\textbf{0.07}$\pm$0.01  & \scriptsize$\alpha\in(2.55, 2.72)$ \newline $\beta\in(9.20,9.95)$
&0.13$\pm$0.05 & \scriptsize$\alpha\in(1.29, 1.34)$ \newline $\beta\in(3.23, 3.41)$
&\textbf{0.19}$\pm$0.00& \scriptsize $\alpha\in(3.20, 3.28)$ \newline $\beta\in (6.51,6.72)$
\vspace{0.01in}\\
GIG & \textbf{0.07}$\pm$0.01 & \scriptsize $\nu\in(-2.61,-2.37)$\newline$a\in(0.01,0.02)$\newline $b\in(17.41,19.14)$
& \textbf{0.12}$\pm$0.01 & \scriptsize$\nu\in(-1.33,-1.28)$\newline$a\in(0.00,0.00)$\newline $b\in(6.42, 6.75)$
 &\textbf{0.19}$\pm$0.00  & \scriptsize$\nu\in(-3.25,-3.18)$\newline$a\in(0.00,0.00)$\newline $b\in(12.93,13.30)$\vspace{0.01in}\\
GGP  &0.15$\pm$0.06 & \scriptsize$\sigma \in(-0.93, -0.80)$ \newline $\tau\in(75.81,85.52)$
& 0.18$\pm$0.02 &  \scriptsize$\sigma \in(0.19, 0.22)$ \newline $\tau\in(11.53,12.98)$
& 0.40$\pm$0.10  & \scriptsize$\sigma \in(-0.18, -0.04)$ \newline $\tau\in(92.05,196.17)$ \\\bottomrule
\end{tabular}
\label{tab:rank1_real}
\end{table}

\subsection{Experiments with latent overlapping communities}
Finally, we tested our models with latent overlapping communities on two real-world graphs
with ground-truth communities.\\
$\bullet$ \texttt{polblogs}\footnote{\url{http://www.cise.ufl.edu/research/sparse/matrices/Newman/polblogs}}: the network of Americal political blogs. 1,224 nodes and 16.715 edges, two true communities (left or right).\\
$\bullet$ \texttt{DBLP}\footnote{\url{https://snap.stanford.edu/data/com-DBLP.html}}: Co-authorship network of DBLP computer science bibliography. The original network has 317,080 nodes. Based on the ground-truth communities extracted in \citet{Yang2012}, we took three largest communities and subsampled 1,990 nodes among them.  The subsampled graph contains 4,413 edges.

\begin{figure}
\centering
\subfigure[\texttt{polblogs}]{\includegraphics[width=0.4\textwidth]{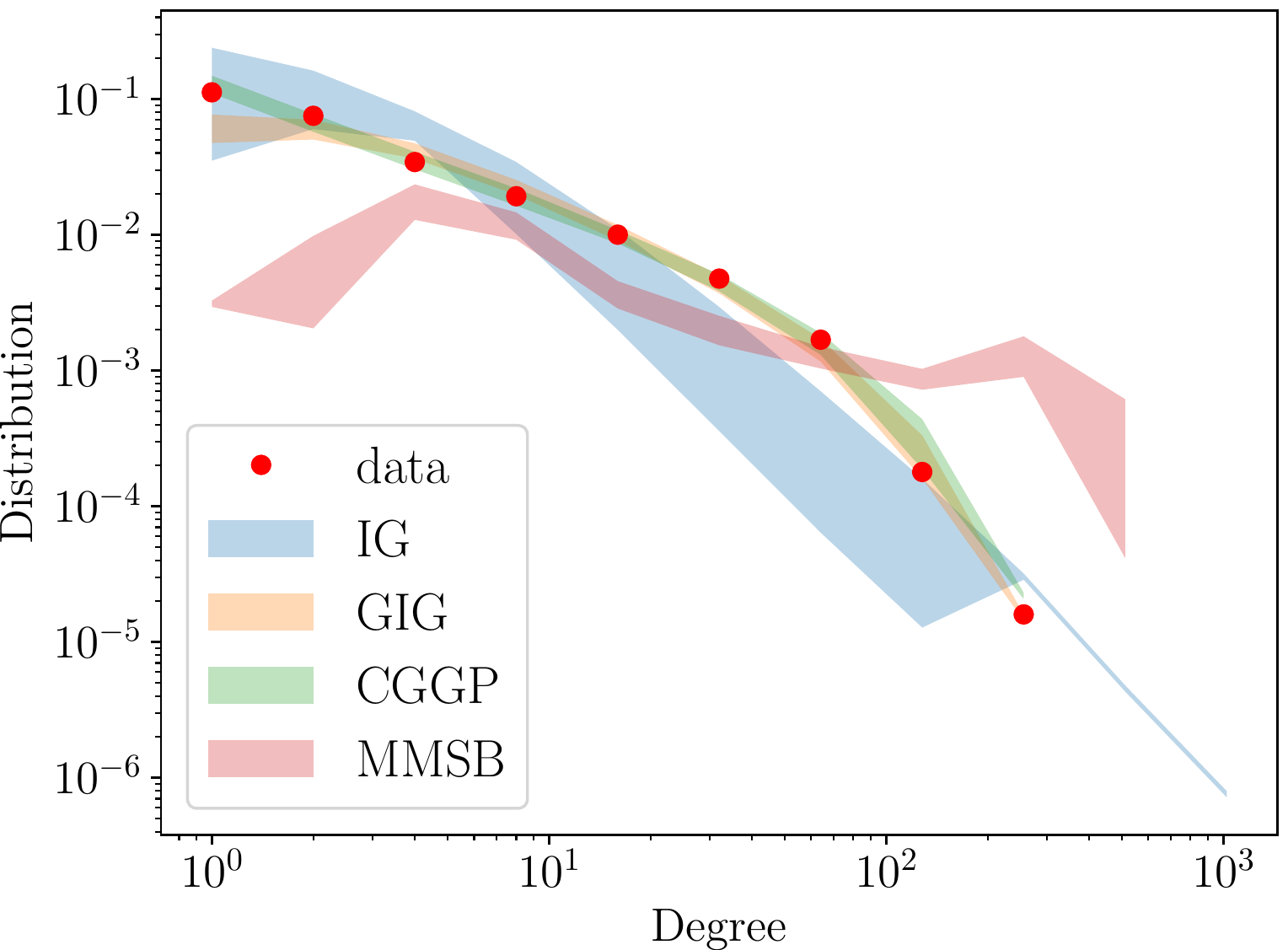}}
\subfigure[\texttt{DBLP}]{\includegraphics[width=0.4\textwidth]{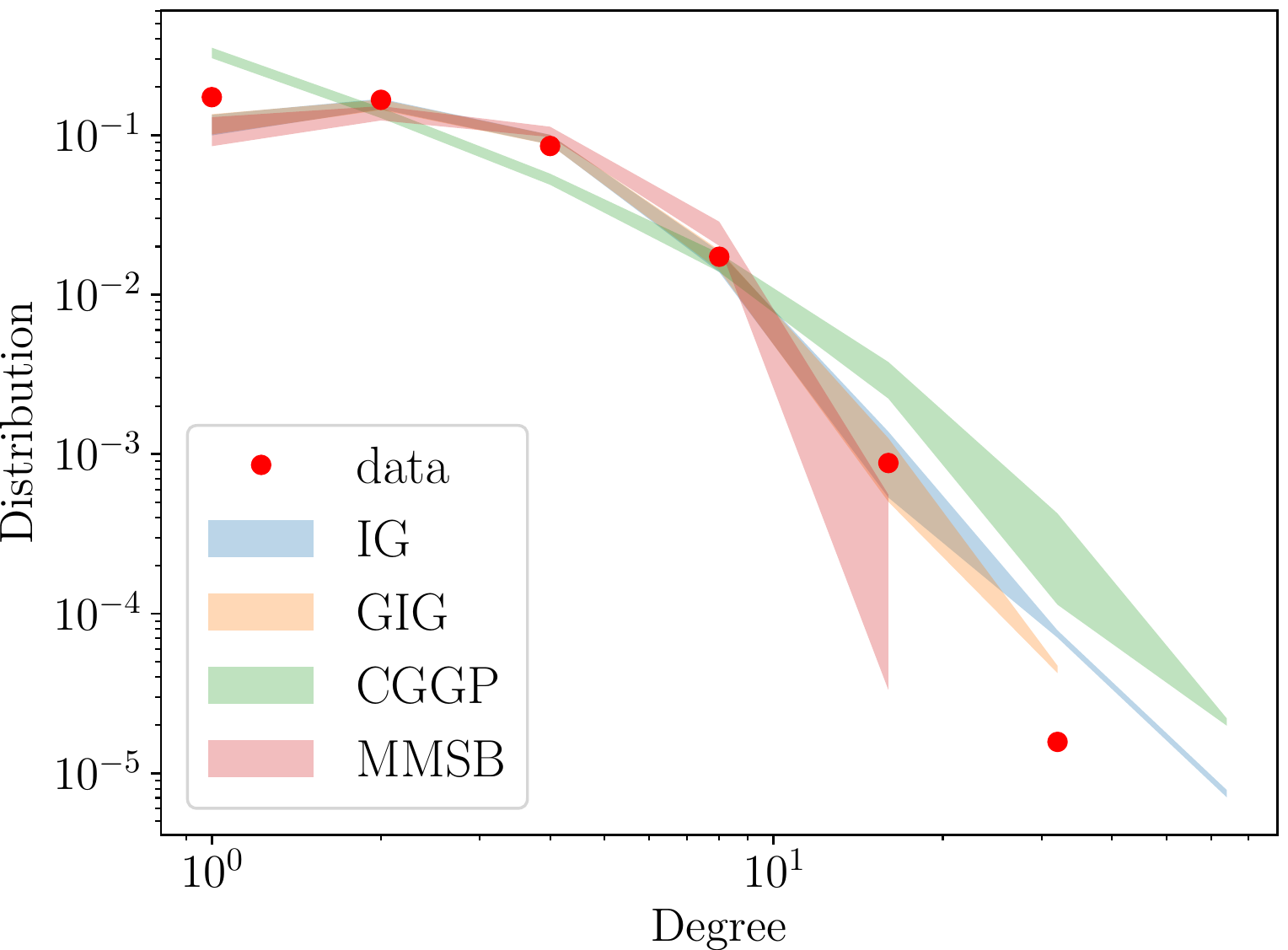}}
\caption{95\%  credible intervals of predictive degree distributions on (a) \texttt{polblogs}
and (b) \texttt{DBLP}. }
\label{fig:rankc_real}
\end{figure}

We compared our two models IG-NR and GIG-NR models to the random graph model based on compound generalized gamma process (CGGP,~\citep{Todeschini2016}), and mixed membership stochastic blockmodel (MMSB,~\cite{Airoldi2009}). CGGP can capture the latent overlapping communities and has asymptotic power-law degree distsribution of exponent in $(1,2)$. MMSB can capture the latent communities, but does not include a degree correction term. For all three models, we set the number of communities to be equal to two for \texttt{polblogs}, and three for \texttt{DBLP}. The CGGP was ran for 200,000 iterations after 10,000 initial iterations where $w$ was initialized by running the model without communities (GGP). Each iteration of the sampler for MMSB scales quadratically with the number of nodes, and the sampler was therefore ran for a smaller number of iterations (5,000) for fair comparison. We found that longer iterations did not lead to improved performances. All methods were ran with three MCMC chains.  For CGGP and MMSB methods, point estimates of the parameters measuring the level of affiliation of each individual  were obtained using the Bayesian estimator described in \citet{Todeschini2016}.
For IG-NR and GIG-IR, we simply took the maximum a posteriori estimate of $V$. To compare to the ground truth communities, nodes are then assigned to the community where they have the strongest affiliation. The learned communities are shown in the appendix. Posterior predictive of the degree distributions for the different models are given in \cref{fig:rankc_real}, and the KS statistic in \cref{tab:KS2}. Both GIG-NR and CGGP exhibit a good fit to the \texttt{polblogs} dataset, where there does not seem to be evidence for a power-law exponent greater than 2. For the DBLP, both IG-NR and GIG-NR provide a good fit, while CGGP fails to capture adequately the degree distribution. The classification accuracy is also reported in \cref{tab:KS2}. The classification accuracy is similar for IG-NR, GIG-NR and CGGP on  \texttt{polblogs}.  IG-NR and GIG-NR outperform other methods on the \texttt{DBLP} network. MMSB failed to capture both degree distributions and community structures, due to the large degree heterogeneity, a limitation already reported in previous articles~\citep{Karrer2011, Gopalan2013}.

\begin{table}[h]
\centering
\caption{Average reweighted KS statistics and clustering accuracies.}
\small
\begin{tabular}{@{}  c  c c c c  @{}}\toprule
& \multicolumn{2}{c}{\texttt{polblogs}} & \multicolumn{2}{c}{\texttt{DBLP}} \\
\cmidrule{2-3} \cmidrule{4-5}
&$D$ & Acc (\%) &$D$ & Acc (\%) \\ \midrule
IG & 0.71$\pm$0.50 & \textbf{94.28} & \textbf{0.08}$\pm$0.03 & 72.46 \\
GIG & 0.14 $\pm$ 0.03 & 93.79  & 0.09$\pm$0.03& \textbf{76.58}\\
CGGP & \textbf{0.12}$\pm$0.03 & 94.12 &0.33$\pm$0.02 & 57.49\\
MMSB & 3.74$\pm$1.18 & 52.12 & 0.37$\pm$0.07 & 39.94
\\\bottomrule
\end{tabular}
\label{tab:KS2}
\end{table}

\bibliographystyle{plainnat}
\bibliography{paper}

\appendix

\clearpage
\begin{appendices}

\section{On scale-free networks}
A recent article by \cite{Broido2018} (BC) raised concerns about the claim that many real-world networks are scale-free. BC performed a statistical analysis on a large number of networks to test whether the degree distribution follows a power-law distribution or some alternative distributions. One of the alternative degree distribution considered is a power-law distribution with an exponential cut-off, which was shown to provide a better fit for a majority of the datasets considered. BC conclude in their article that scale-free networks are rare.

While we agree with the authors that there is a need for rigorous statistical testing of the scale-free hypothesis, and that scale-free networks may indeed by more rare than originally thought, we do not think that the conclusion of the authors is supported by their experiments, except if one considers a very narrow definition of a scale-free network. As pointed out by \cite{Barabasi2018} in a blog post discussing their article, scale-freeness is an asymptotic property: as the sample size goes to infinity, the degree distribution converges to a power-law (up to a slowly varying function, see Definition \ref{def:scalefree}). Degree distributions of finite-size graphs may still depart significantly from a pure power-law distribution.

\begin{figure}[h]
\centering
\includegraphics[width=0.3\linewidth]{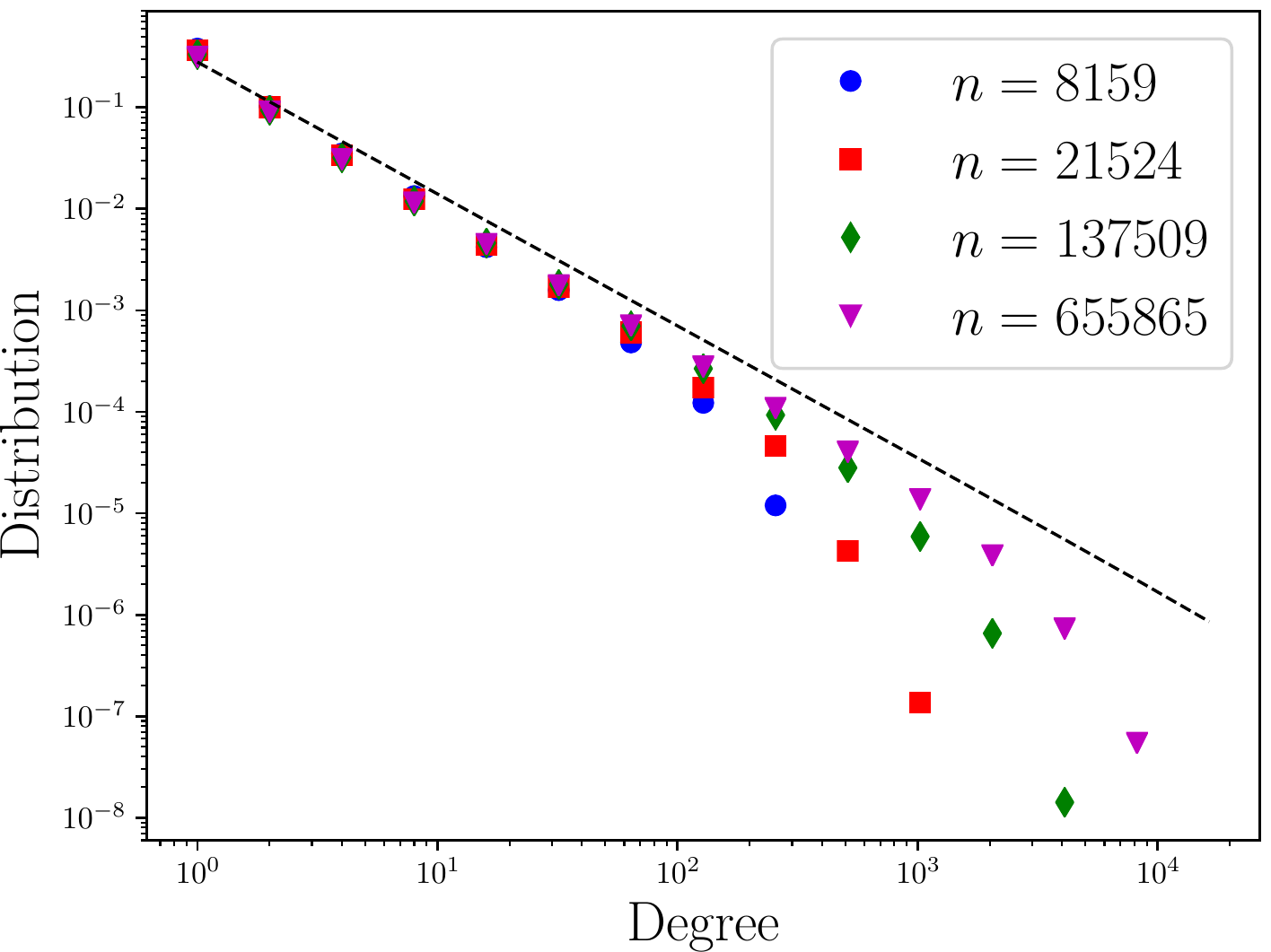}
\includegraphics[width=0.3\linewidth]{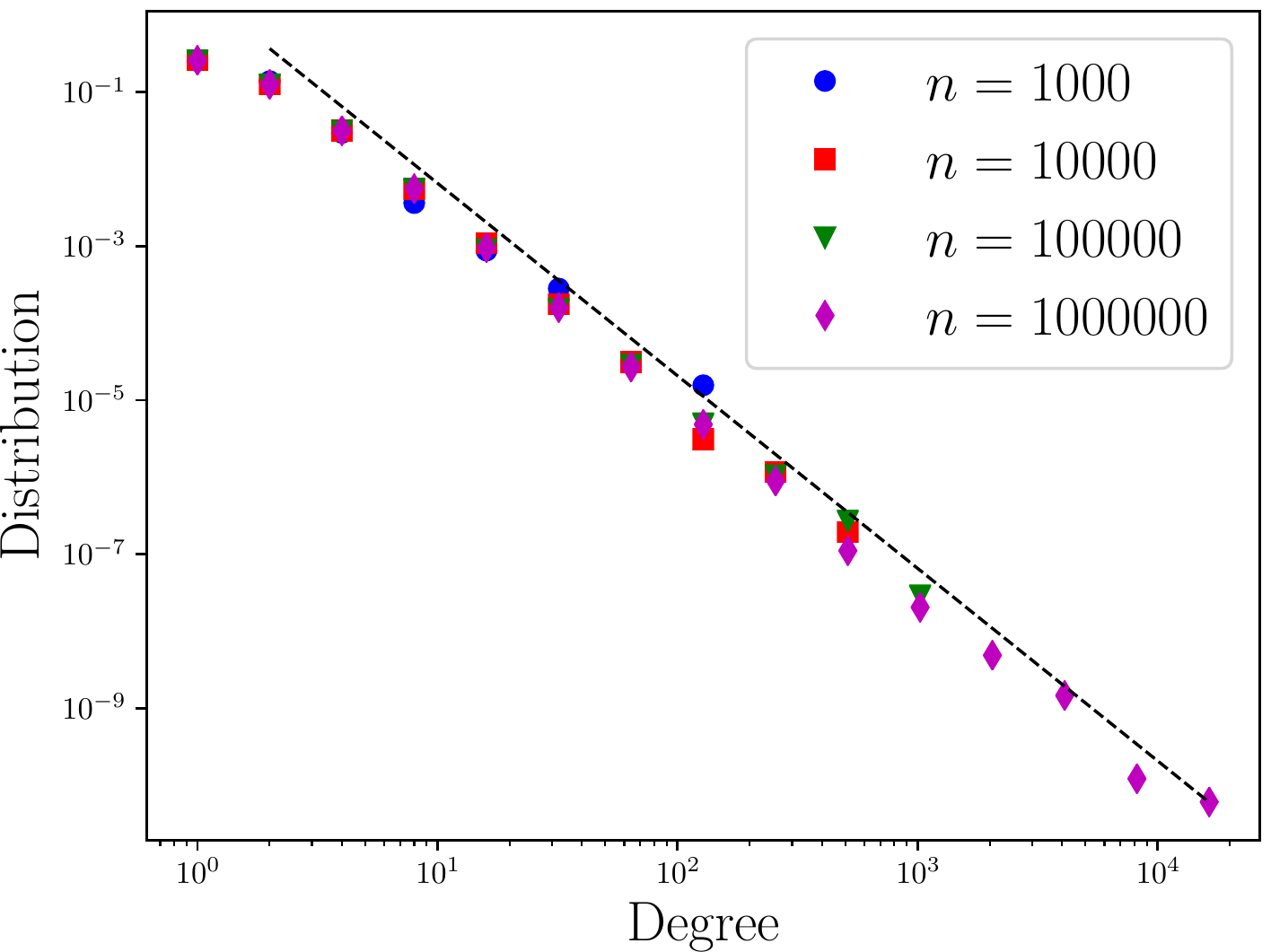}
\includegraphics[width=0.3\linewidth]{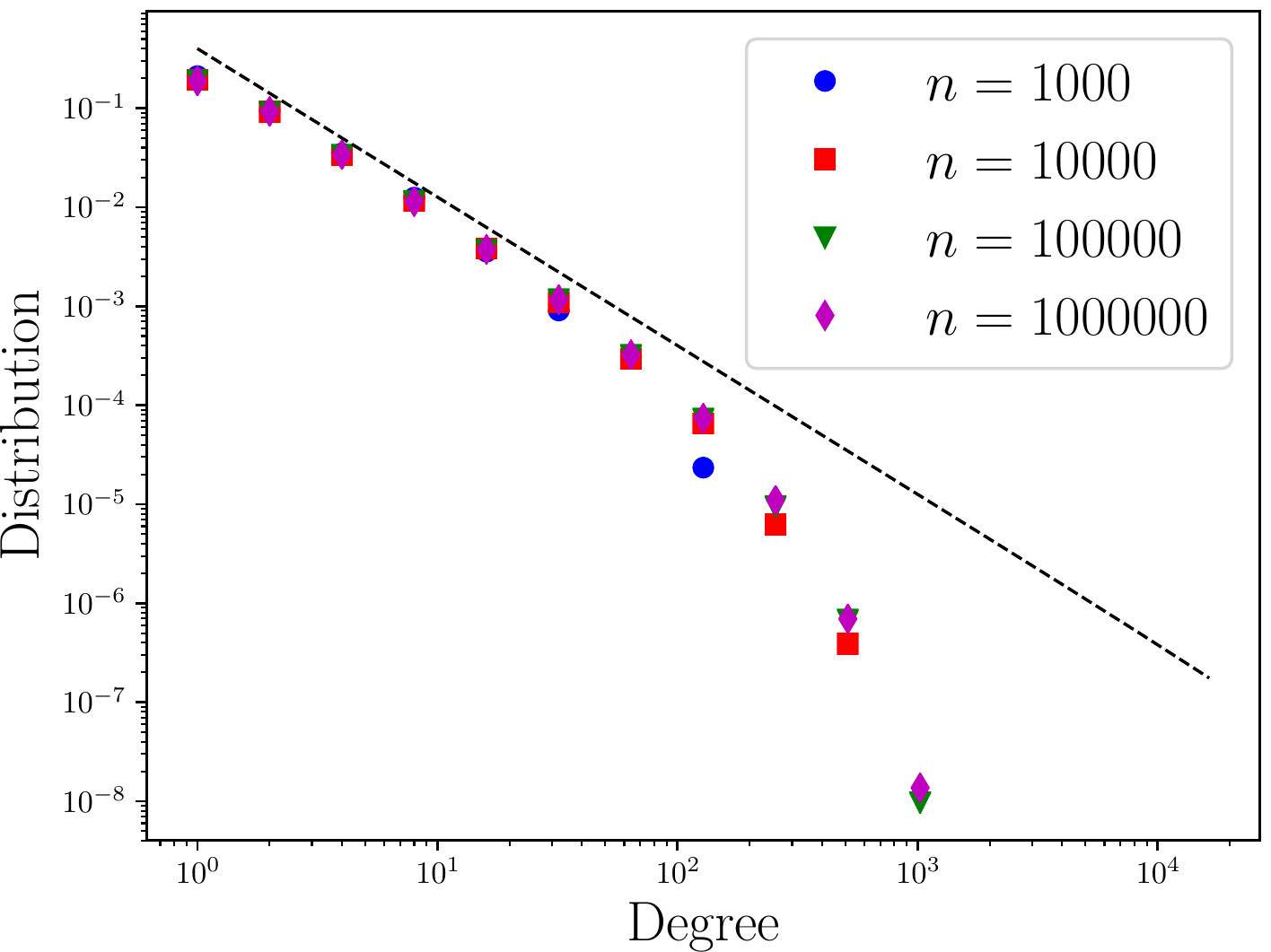}
\caption{Empirical degree distribution for networks for growing sizes generated from the (left) GGP model, (middle) IG-NR model and (right) GIG-NR model. Both the GGP and IG-NR models are scale-free networks, with asymptotically power-law degree distributions.}
\label{fig:scalefree}
\end{figure}

A salient example is given by the class of networks introduced by \cite{Caron2017}, which are known to be scale-free with exponent between 1 and 2 for some values of the parameters. As shown in Figure~\ref{fig:scalefree}, while the degree distribution is asymptotically power-law, any finite-size graph exhibits an exponential cut-off, which shifts to the right as the sample size increases. Therefore, any statistical test on a fixed-n graph is likely to reject the pure power-law hypothesis although the network model is indeed scale-free. 

For reference, empirical degree distributions for the IG-NR and GIG-NR are also plotted in Figure~\ref{fig:scalefree}. The IG-NR is scale-free, and each finite-$n$ distribution is additionally close to a pure power-law distribution. The GIG-NR is not scale-free, and the asymptotic degree distribution is a power-law distribution with exponential cut-off.

\section{Background material on regular variation}
In this section we give some definitions and properties of slowly and regularly varying functions, see the books of \citet{Bingham1989}, \citet{Mikosch1999} and \citet{Resnick2007} for reference.
\begin{defn}
A positive function $f:\mathbb R_+\rightarrow\mathbb R_+$ is regularly varying at infinity with index $\alpha\in\mathbb R$ if $$
\lim_{t\rightarrow\infty}\frac{f(ct)}{f(t)}=c^\alpha$$ for all $c>0$.
\label{def:regvar}
\end{defn}
If $\alpha=0$, the function is said to be slowly varying. Examples of slowly varying include constant functions, functions converging to a constant, logarithms, etc. If a function $f$ is regularly varying with index $\alpha$, then there exists a slowly varying function $\ell$ such that $$f(x)\sim \ell(x)x^\alpha$$ as $x\rightarrow\infty$.

\begin{defn}
 A non-negative random variable\ $X$ with cdf $F_{X}$ is said to be regularly varying
with exponent $\alpha\geq0$ if
\[
1-F_{X}(x)\sim\ell(x)x^{-\alpha}%
\]
as $x$ tends to infinity, where $\ell$ is a slowly varying function at
infinity. If $\alpha>0$ and $F_{X}$ is absolutely continuous with density
$f_{X}$, where $f_{X}$ is ultimately monotone, then
\[
f_{X}(x)\sim\alpha\ell(x)x^{-\alpha-1}%
\]
as $x\rightarrow\infty$.
\label{def:regvarrandomvariable}
\end{defn}

\begin{prop}
Let $X$ be a regularly varying random variable with exponent $\alpha>0$, and
$Y$ be a positive random variable, independent of $X$, with $\mathbb E(Y^{\alpha
+\varepsilon})<\infty$ for some $\varepsilon>0$.\ Then $Z=XY$ is regularly
varying with exponent $\alpha>0$ and
\[
1-F_{Z}(z)\sim \mathbb E(Y^{\alpha})\ell(z)z^{-\alpha}%
\]
as $z$ tends to infinity. If $F_Z$ is absolutely continuous with ultimately monotone density $f_Z$, this implies
$$
f_Z(z)\sim \alpha \mathbb E(Y^{\alpha})\ell(z)z^{-\alpha-1}
$$
as $z$ tends to infinity.
\label{def:propregvar}
\end{prop}

\begin{prop}
If $f$ is a regularly varying function with index $\alpha$, then
$$
\lim_{x\rightarrow\infty}\frac{ \log f(x)}{\log x}=\alpha.
$$
\end{prop}

\section{Background on inhomogeneous random graph models}
\label{sec:bollobas}
In this section, we review the general framework of inhomogeneous random graphs (IRG) presented in \citet{Bollobas2007}.
We start by introducing a \emph{vertex space} and a \emph{kernel} to define IRGs.
\begin{defn}\label{def:vertex_space}
A \textbf{vertex space} $\calV$ is a triplet $(\calS, \mu, (\bx_n)_{n\geq 1})$, where $\calS$ is a separable metric space, $\mu$ is a Borel probability measure on $\calS$, and $\bx_n \defas (x_{1}, \dots, x_{n})$\footnote{To be precise, we should write $\bx_n = (x_1^{(n)}, \dots, x^{(n)}_n)$, but we omit the superscript for simplicity.}
 is a random sequence of $n$ points in $\calS$ such that for each $n\geq 1$,
\[
\nu_n \defas \frac{1}{n} \sum_{i=1}^n \delta_{x_{i}} \pto \mu,
\]
where $\pto$ denotes the convergence in probability. The pair $(\calS, \mu)$ is called a \textbf{ground space}.
\end{defn}

\begin{defn}\label{def:kernel}
A \textbf{kernel} $\kappa$ on a ground space $(\calS, \mu)$ is a symmetric non-negative Borel measurable function on $\calS\times\calS$.
\end{defn}
Rougly speaking, a vertex space is a space of values assigned to vertices in a graph,
such as vertex weights or popularities. Each vertex is associated with a point in $\calS$,
and these points are used to contruct edge probabilities between vertices through the kernel $\kappa$. Kernels should be further restricted to be in a class of functions satisfying
some conditions, and we will explain those shortly after. Given a vertex space and a kernel,
an IRG is defined with link function (edge probabilities)
\[\label{eq:chunglu}
p_{ij}^{(n)} = \frac{\kappa(x_i, x_j)}{n} \wedge 1.
\]
All the following arguments will be explained with this choice of link function,
but everything still holds with the following alternative choices of link functions~\citep[Remark 2.4]{Bollobas2007}.
\[
p_{ij}^{(n)} &= 1 - \exp\bigg(-\frac{\kappa(x_i,x_j)}{n}\bigg)\label{eq:nr}\\
p_{ij}^{(n)} &= \frac{\kappa(x_i,x_j)}{n + \kappa(x_i,x_j)}\label{eq:grg}.
\]
All these three functions are related to existing works on IRGs. The link function \eqref{eq:chunglu}
is a generic version of \cite{Chung2002}, \eqref{eq:nr} is for \cite{Norros2006}, and \eqref{eq:grg} is for \cite{Britton2006}. We chose \eqref{eq:nr} for our model
because of the computational efficiency in posterior inference.

Let $G_n$ be a graphs generated from IRG described above with a vertex space $\calV$ and
a kernel $\kappa$. The kernel $\kappa$ are assumed to be \emph{graphical},
which is defined as follows.
\begin{defn}\label{def:graphical_kernel}
A kernel $\kappa$ on a vertex space $(\calS, \mu, (\bx_n))$ is \textbf{graphical} if the followings hold:
\begin{enumerate}[(i)]
\item $\kappa$ is continuous almost everywhere on $\calS\times \calS$.
\item $\kappa \in L^1(\calS\times \calS, \mu\times\mu)$.
\item Let $E_n$ be the set of edges in $G_n$. Then,
\[
\lim_{n\to\infty}\frac{1}{n} \bbE(|E_n|) = \frac{1}{2}\iint_{\calS\times\calS}\kappa(x, y)d \mu(x) d\mu(y).
\]
\end{enumerate}
\end{defn}
The first and second conditions are natural technical requirements. The third condition is related to the density of graphs. It requires $\kappa$ to measure the density of the edges~\citep{Bollobas2007}.

The following theorem characterizes the asymptotic degree distribution of IRGs.
\begin{thm}
\label{thm:bollobas_degree}
(\citep[Theorem 3.13]{Bollobas2007}) Let $\kappa$ be a graphical kernel on a vertex space $\calV$. For any fixed $k\geq 0$,
\[
\frac{N_k^{(n)}}{n} \pto \int_\calS \frac{\lambda(x)^k}{k!}e^{-\lambda(x)} d\mu(x),
\]
where $N_k^{(n)}$ is the number of vertices in with degree $k$ in $G_n$, and
\[
\lambda(x) \defas \int_\calS \kappa(x, y) d\mu(y).
\]
\end{thm}
Hence, one can easily compute the asymptotic degree distribution of any IRG that fits into the framework once the corresponding vertex space and kernel is specified. This is what we do in the next two sections.

\section{Proof of \cref{thm:rank1_sparsity} and \cref{thm:rank1_degree}}
\cref{thm:rank1_sparsity} and \cref{thm:rank1_degree} in the main paper are directly obtained by showing that the Norros-Reittu IRG (NR-IRG) fits into the general framework discussed in \cref{sec:bollobas} of this supplementary material. Actually, the NR-IRG has been discussed as an example of rank-1 IRGs, see~\citep[Section 16.4]{Bollobas2007}. More precisely, define a vertex space $\calV = (\calS, \mu, \bx_n)$ with
\[
\calS = (0,\infty), \quad
\mu = \calL_{w_1}, \quad
x_{i} = w_i \sqrt{\frac{\bbE(w_1)}{s\sn/n}},
\]
where $\calL_{w_1}$ denotes the law of $w_1$, and define a kernel
\[
\kappa(x, y) = \frac{xy}{\bbE(w_1)}.
\]
To see if this kernel is graphical, note that
\[
\lefteqn{\lim_{n\to\infty} \frac{1}{n}\bbE(|E_n|)} \nonumber\\
& =
\lim_{n\to\infty} \frac{1}{n}\bbE \Bigg[
 \sum_{i<j} \bigg\{1 - \exp\bigg( - \frac{\kappa(x_i,x_j)}{n}\bigg)\bigg\} \Bigg]\nonumber\\
 &= \lim_{n\to\infty} \frac{1}{n} \Bigg[
 \sum_{i<j} \bigg\{1 - \exp\bigg( -\frac{w_iw_j}{s\sn}\bigg)\bigg\} \Bigg]\nonumber\\
 &\leq  \lim_{n\to\infty} \frac{1}{n} \bbE\Bigg[ \sum_{i<j} \frac{w_iw_j}{s\sn}\Bigg] \nonumber\\
 &\leq \lim_{n\to\infty} \frac{1}{2} \bbE(s\sn/n) = \bbE(w_1)/2 \nonumber\\
 &= \frac{1}{2} \iint_{\calS^2} \kappa(x, y) d\mu(x) d\mu(y).
\]
Hence, combined with \citet[Lemma 8.1]{Bollobas2007}, we get
\[
\lim_{n\to\infty} \frac{1}{n}\bbE(|E_n|) = \frac{1}{2}\iint_{\calS^2} \kappa(x, y) d\mu(x) d\mu(y).
\]
and the kernel $\kappa$ is therefore graphical. The second part of \cref{thm:rank1_sparsity} then follows from \citet[Proposition 8.9]{Bollobas2007}, and \cref{thm:rank1_degree} follows from  \cref{thm:bollobas_degree} with
\[
\lambda(x) = \int_\calS \kappa(x, y) d\mu(y)
= \int_0^\infty \frac{xy}{\bbE(w_1)} \dee \calL_{w_1}(y) = x.
\]

\section{Proof of \cref{thm:rankc_sparsity_and_degree}}
\cref{thm:rankc_sparsity_and_degree} also follows by showing that the rank-$c$ model fits into the general framework discussed in \cref{sec:bollobas}. Define a vertex space $\calV = (\calS, \mu, \bx_n)$ with
\[
\calS = (0, \infty)^{c+1},\quad \mu = \calL_{w_1} \calL_{(v_{11}, \dots, v_{1c})},
\]
where $\calL_{w_1}$ and $\calL_{(v_{11}, \dots, v_{1c})}$ denote the laws of $w_1$ and $(v_{11},\dots, v_{1c})$, and
\[
x_i = \Bigg( w_i \sqrt{\frac{\bbE(w_1)}{s\sn/n}}, v_{i1}\sqrt{\frac{\bbE(v_{11})}{r_{1}\sn/n}},
\dots, v_{ic} \sqrt{\frac{\bbE(v_{1c})}{r_{c}\sn/n}}\Bigg).
\]
Define a kernel on this space
\[
\kappa(x, y) = \frac{x_{[1]}y_{[1]}}{\bbE(w_1)} \sum_{q=1}^c \frac{x_{[q+1]}y_{[q+1]}}{\bbE(v_{1q})},
\]
where $x_{[d]}$ denotes the $d$th component of $x$. To see if this kernel is graphical, note that
\[
\lefteqn{\lim_{n\to\infty} \frac{1}{n}\bbE(|E_n|)}\nonumber\\
&= \lim_{n\to\infty} \bbE\Bigg[ \frac{1}{n} \sum_{i<j} \bigg\{1 - \exp\bigg(- \frac{w_i w_j}{s\sn}
\sum_{q=1}^c \frac{v_{iq}v_{jq}}{r_{q}\sn/n}\bigg)\bigg\}\Bigg]\nonumber\\
&\leq \lim_{n\to\infty} \bbE \Bigg[ \frac{1}{n} \sum_{i<j} \frac{w_iw_j}{s\sn}
 \bbE\bigg[\sum_{q=1}^c \frac{v_{iq}v_{jq}}{r_{q}\sn/n}\bigg]\Bigg].
\]
Now note that
\[
\bbE\bigg[\sum_{q=1}^c \frac{v_{iq}v_{jq}}{r_{q}\sn/n}\bigg] &=
\frac{1}{n^2}\sum_{i',j'} \bbE\bigg[\sum_{q=1}^c \frac{v_{i'q}v_{j'q}}{r_{q}\sn/n}\bigg]\nonumber\\
&\leq \bbE\bigg[ \sum_{q=1}^c r_{q}\sn/n\bigg] = \sum_{q=1}^c \bbE(v_{1q}).
\]
Plugging this into the above equation yields
\[
\lefteqn{\lim_{n\to\infty} \frac{1}{n}\bbE(|E_n|)}\nonumber\\
&\leq \lim_{n\to \infty} \bbE\Bigg[ \frac{1}{n} \sum_{i<j} \frac{w_iw_j}{s\sn}
\sum_{q=1}^c \bbE(v_{1q})\Bigg] \nonumber\\
&\leq \lim_{n\to\infty} \frac{1}{2} \bbE(s\sn/n) \sum_{q=1}^c \bbE(v_{1q}) \nonumber\\
&= \frac{1}{2}\bbE(w_1)\sum_{q=1}^c \bbE(v_{1q}) \nonumber\\
&= \frac{1}{2} \iint \kappa(x, y) d\mu(x) d\mu(y).
\]
Hence, by  \citet[Lemma 8.1]{Bollobas2007}, we get
\[
\lim_{n\to\infty} \frac{1}{n}\bbE(|E_n|) = \frac{1}{2}\iint_{\calS^2} \kappa(x, y) d\mu(x) d\mu(y).
\]
The second part of \cref{thm:rank1_sparsity} then follows from \citet[Proposition 8.9]{Bollobas2007}, and \cref{thm:rank1_degree} follows from  \cref{thm:bollobas_degree} with
\[
\lambda(x) = \int_\calS \kappa(x, y) d\mu(y) = x_{[1]} \sum_{q=1}^c x_{[q+1]}.
\]

\section{Details on posterior inferences}
\subsection{Posterior inference for the rank-1 model}
The posterior ineference for rank-1 model is summarized in three steps.
\begin{enumerate}
\item Sample $w$ via HMC (we use the transformation $w = e^{\hat w}$ and update $\hat w$).
\item Sample $m$ from truncated Poisson distribution,
\[
p(m_{ij}|w) = \frac{(\frac{w_iw_j}{s\sn})^{m_{ij}}\exp(-\frac{w_iw_j}{s\sn})\indicator{m_{ij}>0}}{m_{ij}!(1-e^{-\frac{w_iw_j}{s\sn}})}.
\]
\item Sample hyperparameters for $p(w)$ via Metropolis-Hastings.
\end{enumerate}
We used the step size $\epsilon=10^{-2}$ and the number of leapfrog steps $L=20$ for all experiments.
\paragraph{Sampling hyperparameters for IG.}
In IG we have two hyperparameters $\alpha > 1$ and $\beta > 0$. We place a log-normal prior on $\alpha-1$ and $\beta$.
\[
\alpha &= 1+e^{\hat\alpha}, \,\, \hat\alpha \sim \calN(0, 1)\\
\beta &= e^{\hat\beta}, \,\, \hat\beta\sim\calN(0, 1).
\]
Then we updated $\hat\alpha$ and $\hat\beta$ via Metropolis-Hastings with proposal distribution $\calN(\hat\alpha, 0.05)$ and $\calN(\hat\beta, 0.05)$. We found that the initialization of $\hat\alpha$ and $\hat\beta$ was important to capture degree distributions. We initialized $\hat\alpha \sim \calN(0, 0.1)$ and using the asymptotic relation
\[
\frac{|E_n|}{n} \pto \frac{\bbE(w_1)}{2} = \frac{\beta}{2(\alpha-1)},
\]
set
\[
\hat\beta = \log \frac{2(\alpha-1)|E_n|}{n}.
\]
\paragraph{Sampling hyperparameters for GIG.}
We have three parameters $\nu<0$, $a>0$, and $b>0$ (we restricted $\nu<0$ to get the positive power-law exponent). We placed log-normal priors on $-\nu, a$ and $b$.
\[
\nu &= -e^{\hat\nu}, \,\, \hat\nu\sim\calN(0,1)\\
a &= e^{\hat a},\,\, \hat a\sim\calN(0,1)\\
b &= e^{\hat b},\,\, \hat b \sim \calN(0,1).
\]
We updated $\hat\nu,\hat a$ and $\hat b$ via Metropolis-Hastings with proposal distributions $\calN(\hat\nu, 0.05), \calN(\hat a, 0.1)$ and $\calN(\hat b, 0.05)$. We initialized $\nu \sim \mathrm{Unif}(-1, 0)$ and $a \sim \mathrm{Unif}(0, 10^{-3})$. $b$ was initialized by solving
\[
\frac{|E_n|}{n} = \frac{\bbE(w_1)}{2} = \frac{\sqrt{b}K_{\nu+1}(\sqrt{ab})}{\sqrt{a}K_\nu(\sqrt{ab})},
\]
using a numerical root-finding algorithm.

\subsection{Posterior inference for the rank$-c$ model}
The posterior inference for the rank$-c$ model is quite similar to that of the rank-1 model.
\begin{enumerate}
\item Sample $w$ via HMC (we use the transformation $w = e^{\hat w}$ and update $\hat w$).
\item Sample $V$ via HMC (see below).
\item Sample $M$ from multivariate truncated Poisson distribution,
\[
p(M_{ij}|w, V) = \prod_{q=1}^c \frac{\lambda_{ijq}^{m_{ijq}} e^{-\lambda_{ijq}}
\indicator{\sum_{ijq'}m_{ijq'}>0}}{1-\exp(-\sum_{q'=1}^c \lambda_{ijq'})},
\]
where $\lambda_{ijq} = \frac{w_iw_j}{s\sn}\frac{v_{iq}v_{jq}}{r_{q}\sn/n}$.
\item Sample hyperparameters of $p(w)$ and $p(V)$.
\end{enumerate}

\paragraph{Details on HMC for $V$.}
Each vector $(v_{i1}, \dots, v_{ic})$ is a Dirichlet random variable such that $\sum_{q=1}^c v_{iq}=1$, so transforming it to an unconstrained vector is quite tricky. We adapt the trick presented in \citet{Betancourt2010}. Let $v = (v_{1}, \dots, v_c) \sim \mathrm{Dir}(\gamma_1, \dots, \gamma_c)$. Define i.i.d. beta random variables,
\[
z_{1} &\sim \mathrm{beta}(\tilde\gamma_1, \gamma_1) \\
z_{2} &\sim \mathrm{beta}(\tilde\gamma_2, \gamma_2) \\
 &\vdots \\
 z_{q-1} &\sim \mathrm{beta}(\tilde\gamma_{q-1}, \gamma_{q-1}),
\]
where
\[
\tilde\gamma_q = \sum_{t=q+1}^c \gamma_t.
\]
Then, if we take a transform
\[
v_{q} = (1 - z_{q}) \prod_{t=1}^{q-1} z_{t} \,\,\, (t < c), \quad
v_{c} = \prod_{t=1}^{c-1} z_{t},
\]
we have $v \sim \mathrm{Dir}(\gamma_1, \dots, \gamma_c)$. The advantage of this transformation is that the Jacobian can be computed efficiently. By the chain rule, we have
\[
\frac{\partial f(v)}{\partial z_{q}} = \frac{v_{q}}{z_{q}-1} \frac{\partial f(v)}{\partial v_{q}} + \sum_{t=q+1}^c \frac{x_t}{z_q} \frac{\partial f(v)}{\partial v_t}.
\]
We take another logistic transform on $z_q$ to make it completely unconstrained.
\[
z_q = \frac{1}{1 + e^{-\hat z_q}},
\]
Hence, the gradient for the unconstrained variable $\hat z_q$ is computed as
\[
\frac{\partial f(v)}{\partial \hat z_q} = \frac{\partial f(v)}{\partial z_q} (z_q - z_q^2).
\]
In our algorithm,  HMC for $V$ is done on the unconstrained variables $(\hat z_{i1}, \dots, \hat z_{i,q-1})_{i=1}^n$.

\paragraph{Initialization and step sizes.}
Unlike \citet{Todeschini2016} where the model is initialized by running MCMC for the simplified model without communities to initialize $w$, we initialize the chain by running MCMC only for $V$ while holding $w$ fixed as $[1,\dots, 1]^\top$. We found this helpful for the algorithm to discover better community structures. For this initialization, we ran HMC for $V$ with $\epsilon=10^{-1}$ and $L=20$. After initialization, we ran HMC for $w$ with $\epsilon=5\cdot 10^{-3}$ and $L=20$, and ran HMC for $V$ with $\epsilon=2.5\cdot 10^{-2}$ and $L=20$. We decayed $\epsilon$ for $V$ to $5\cdot 10^{-3}$ after burn-in.

\paragraph{Sampling hyperparameters for $p(V)$.}
We assume log-normal prior distributions on the hyperparameters $\gamma_1, \dots, \gamma_c$.
\[
\gamma_q = e^{\hat\gamma_c}, \,\,\, \hat\gamma_q \sim \calN(0, 1).
\]
Then we updated $\hat\gamma_q$ via Metropolis-Hastings with proposal distribution $\calN(\hat\gamma_q, 0.01)$. We initialized $\gamma_1= \dots = \gamma_c = 0.1$.

\section{Additional Figures}
\subsection{Empirical degree distributions and number of edges for rank$-c$ model}
\begin{figure}
\centering
\includegraphics[width=0.48\linewidth]{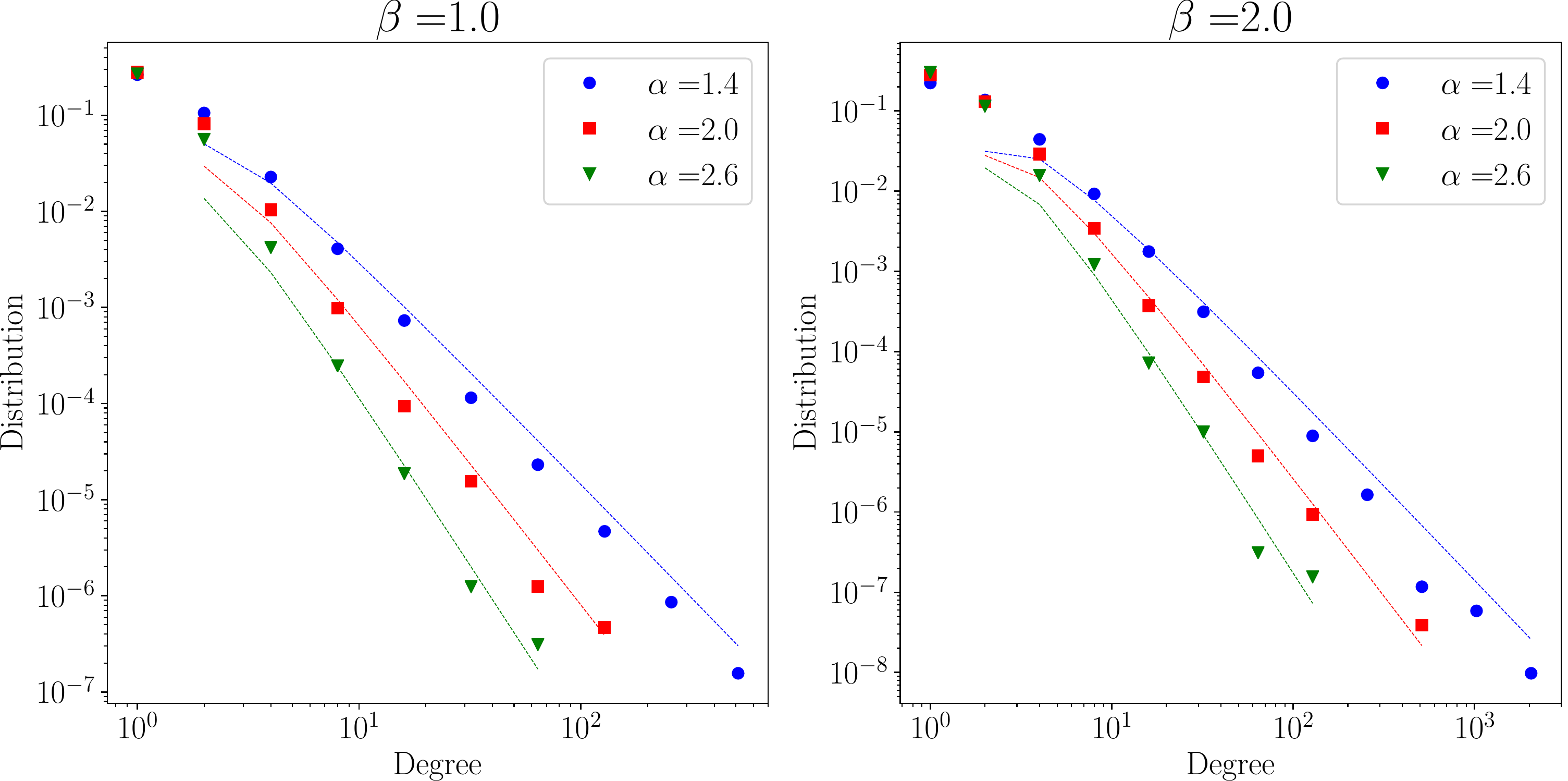}
\includegraphics[width=0.48\linewidth]{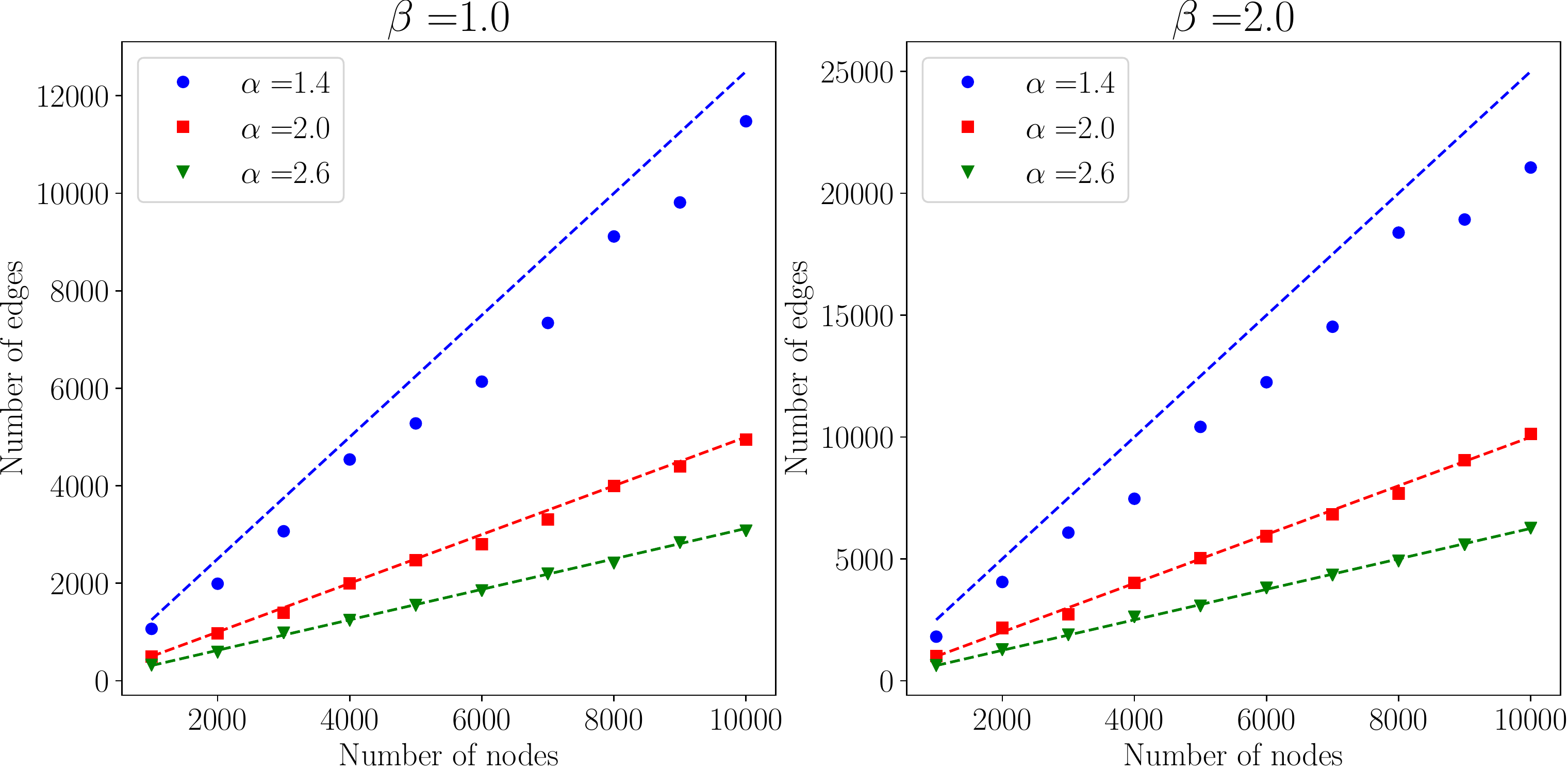}
\includegraphics[width=0.48\linewidth]{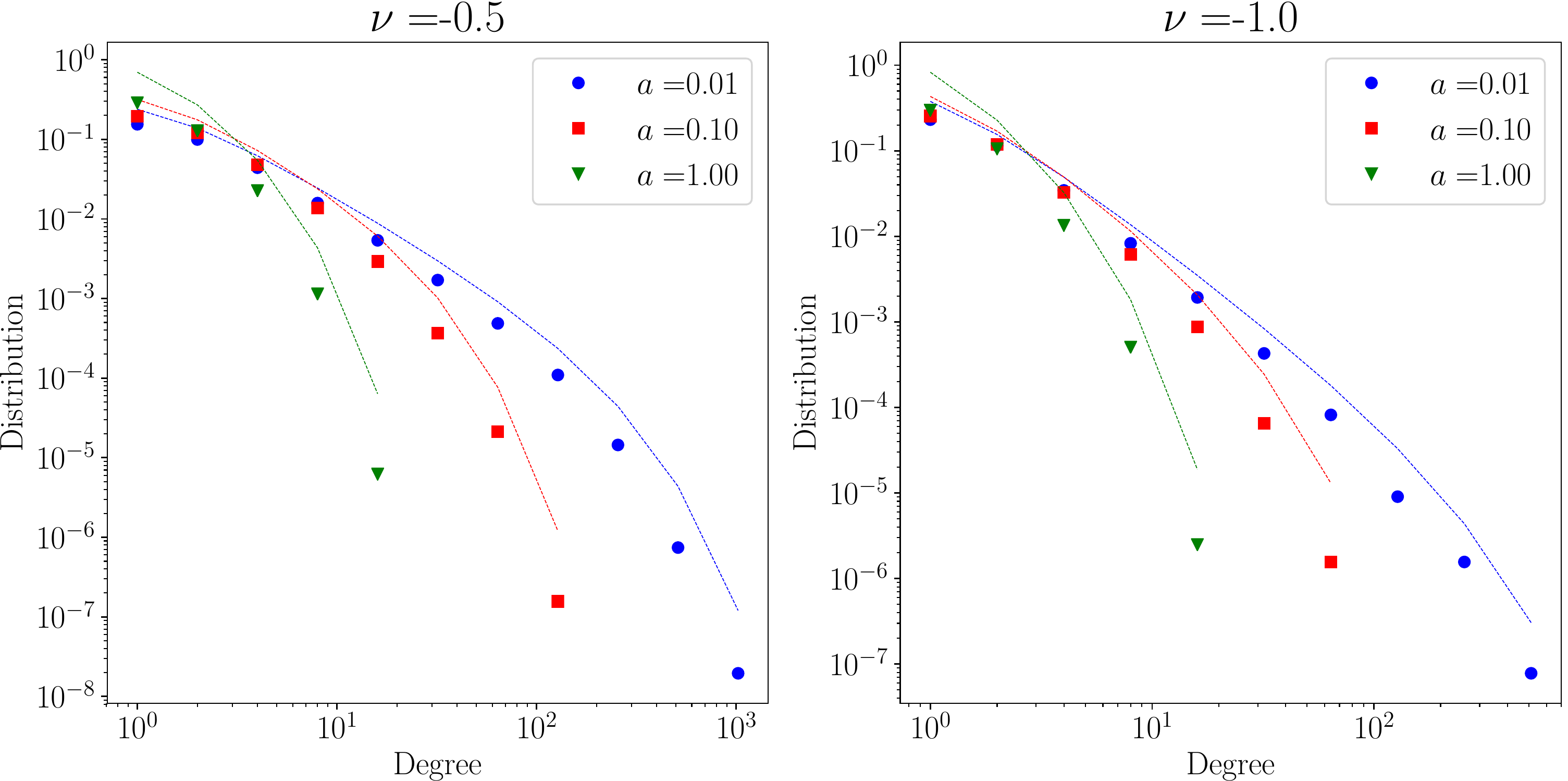}
\includegraphics[width=0.48\linewidth]{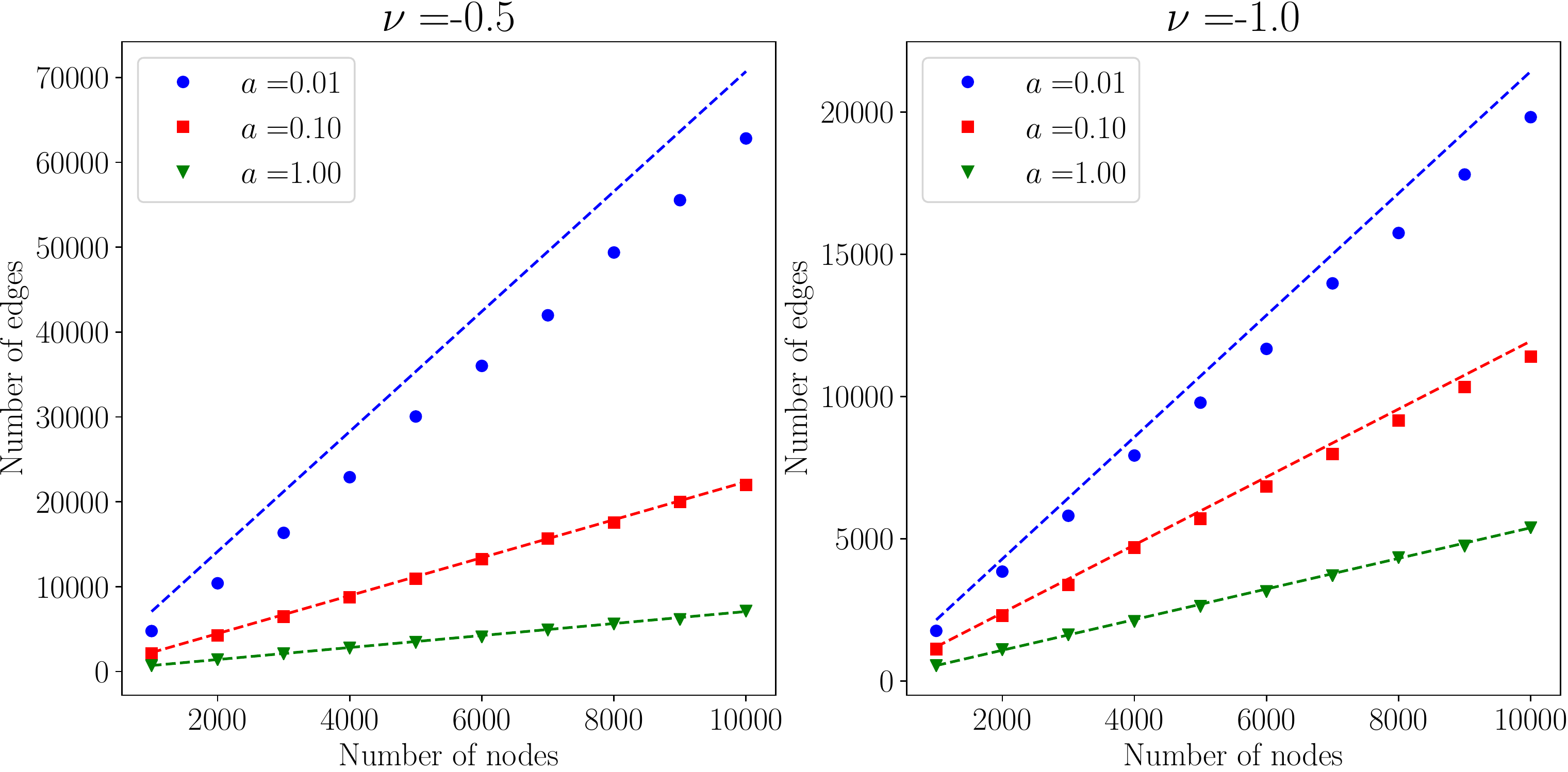}
\caption{
First row, first and second boxes: empirical degree distributions (dashed lines) of graphs with 10,000 nodes sampled from rank-$c$ inverse gamma NR model  compared to the
theoretically expected asymptotic degree distribution (dotted lines), with various values of $\alpha$ and $\beta$. First row, third and fourth boxes: empirical number of edges (dashed lines) of graphs sampled from rank$-c$ inverse gamma NR model versus the number of nodes compared to the theoretically expected value of number of edges (dotted lines), with various values of $\alpha$ and $\beta$. Second row: the same figures for rank$-c$ GIG NR model with various values of $\nu$ and $a$ with fixed $b=2.0$. Best viewed magnified in color.}
\label{fig:rankc_degree_sparsity}
\end{figure}

We first demonstrate the empirical degree distributions and number of edges of graphs generated from rank$-c$ model with $c=5$. The results are presented in \cref{fig:rankc_degree_sparsity}.  As predicted from \cref{thm:rankc_sparsity_and_degree}, the degree distribution and sparsity are not affected by the introduction of the community affiliation factors $V$.

\subsection{Discovered community structures}
\begin{figure}[t]
\centering
\includegraphics[width=0.19\linewidth]{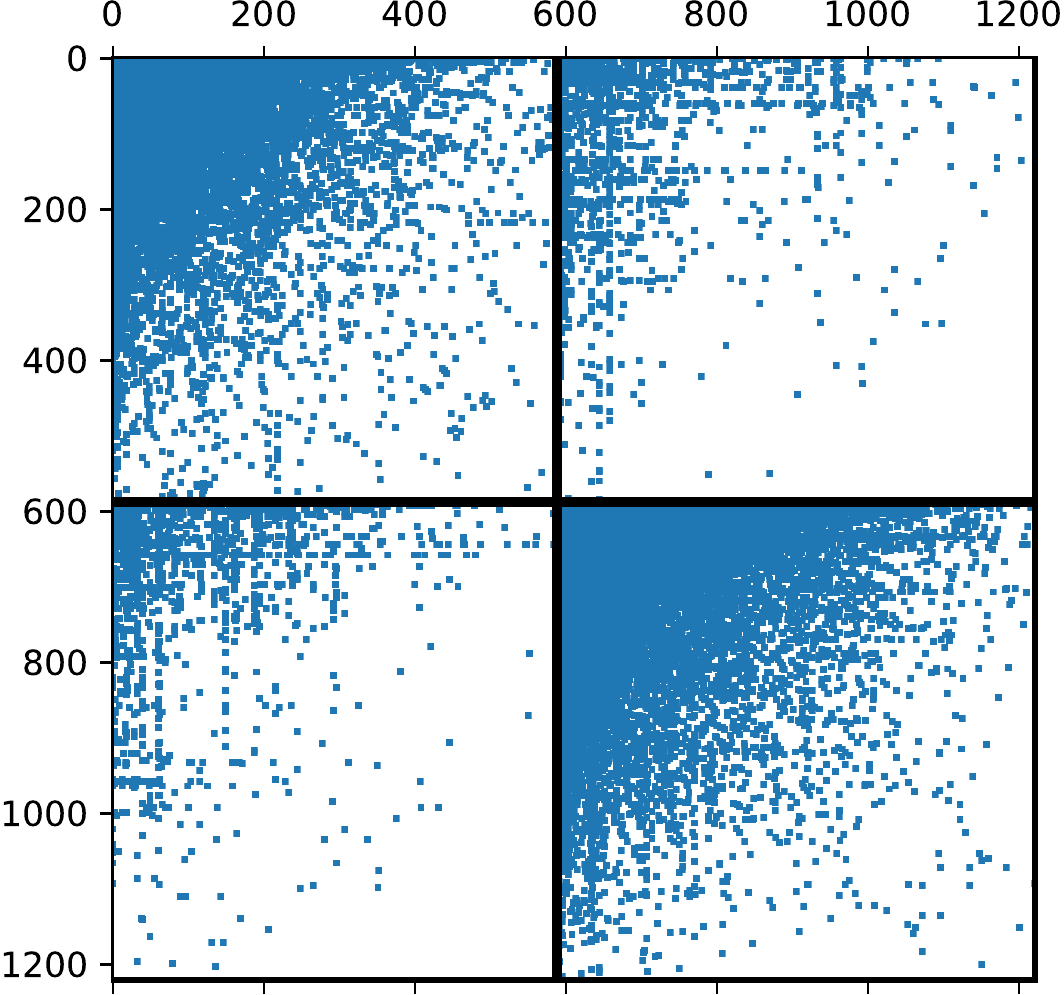}
\includegraphics[width=0.19\linewidth]{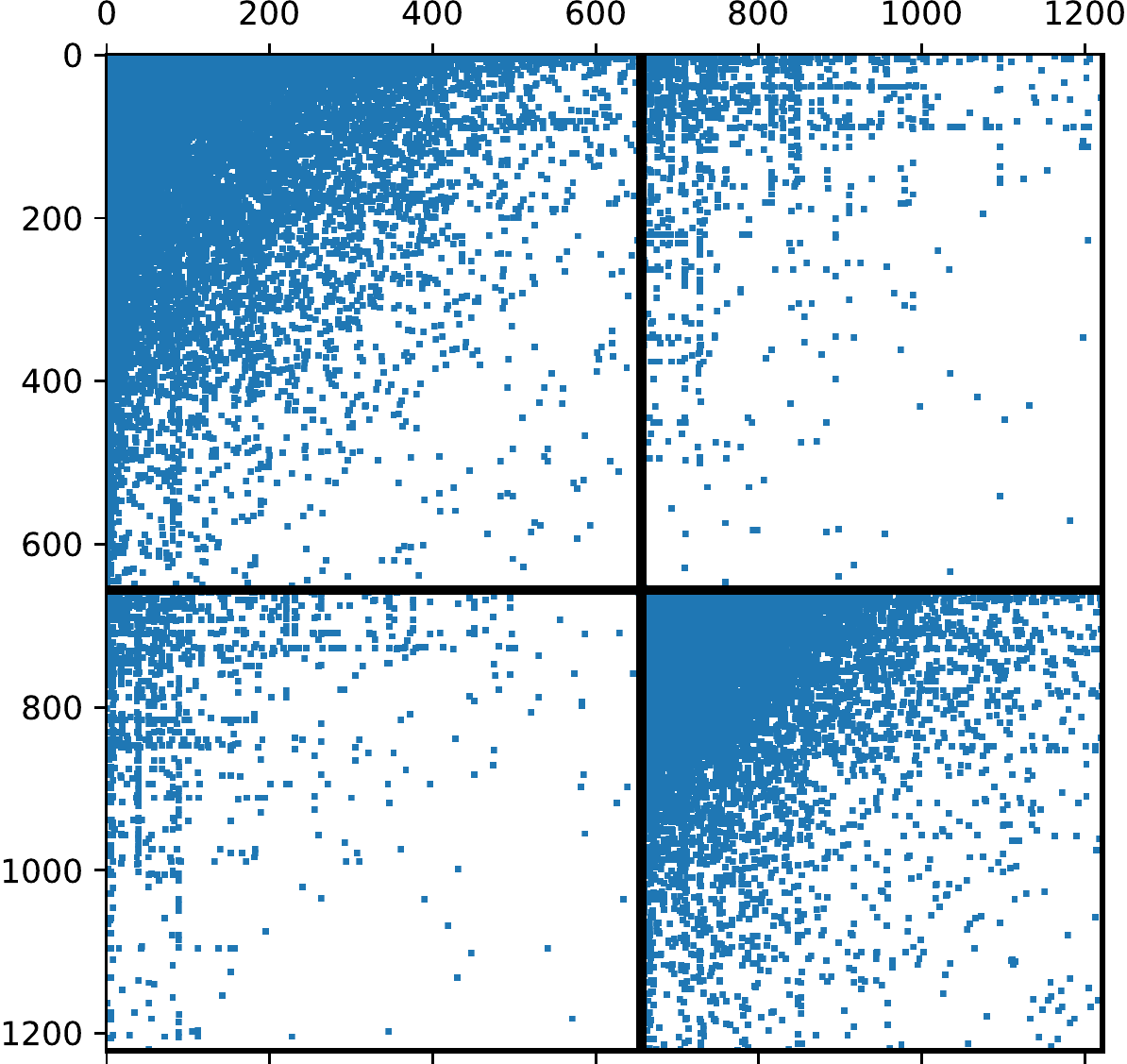}
\includegraphics[width=0.19\linewidth]{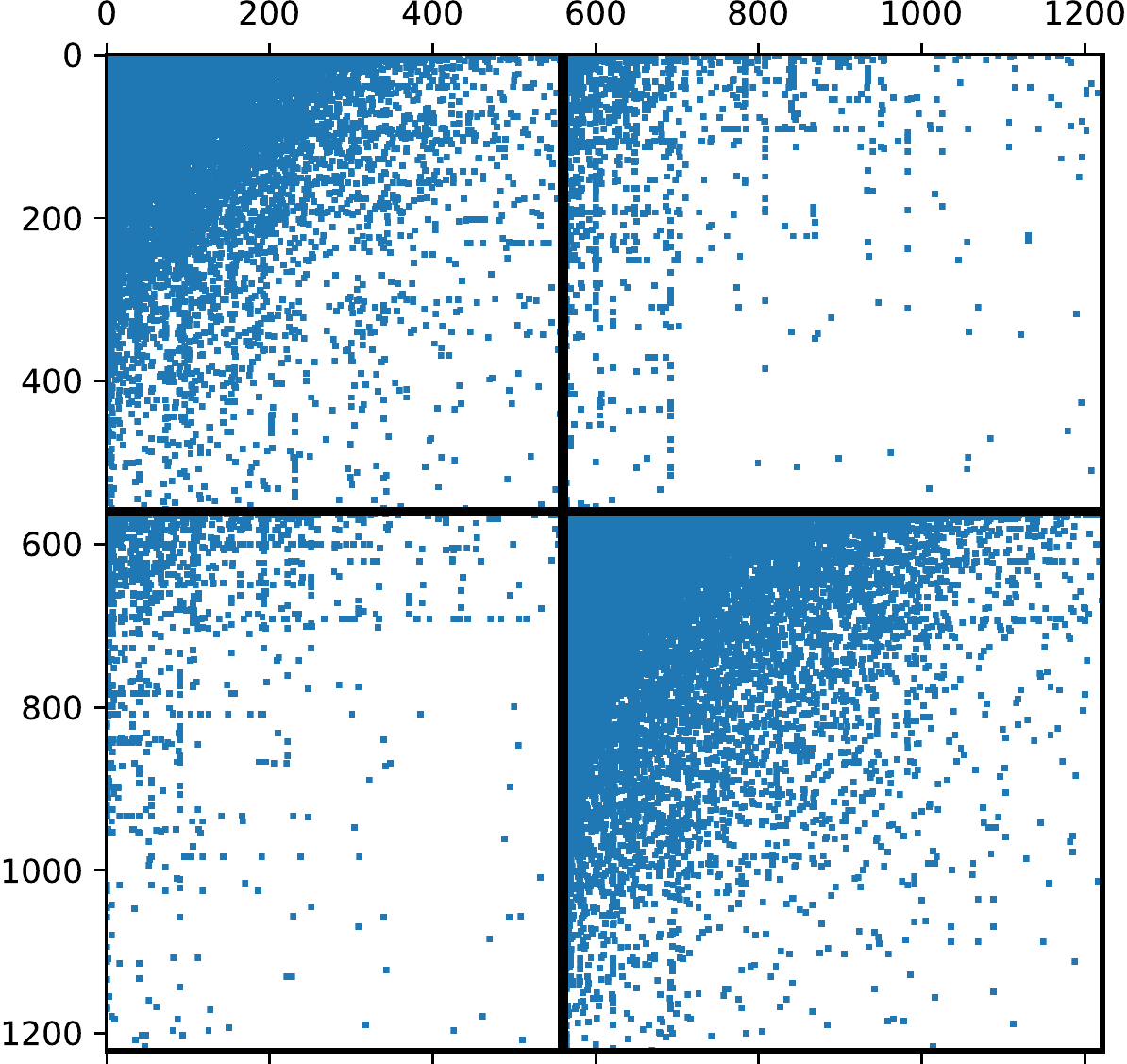}
\includegraphics[width=0.19\linewidth]{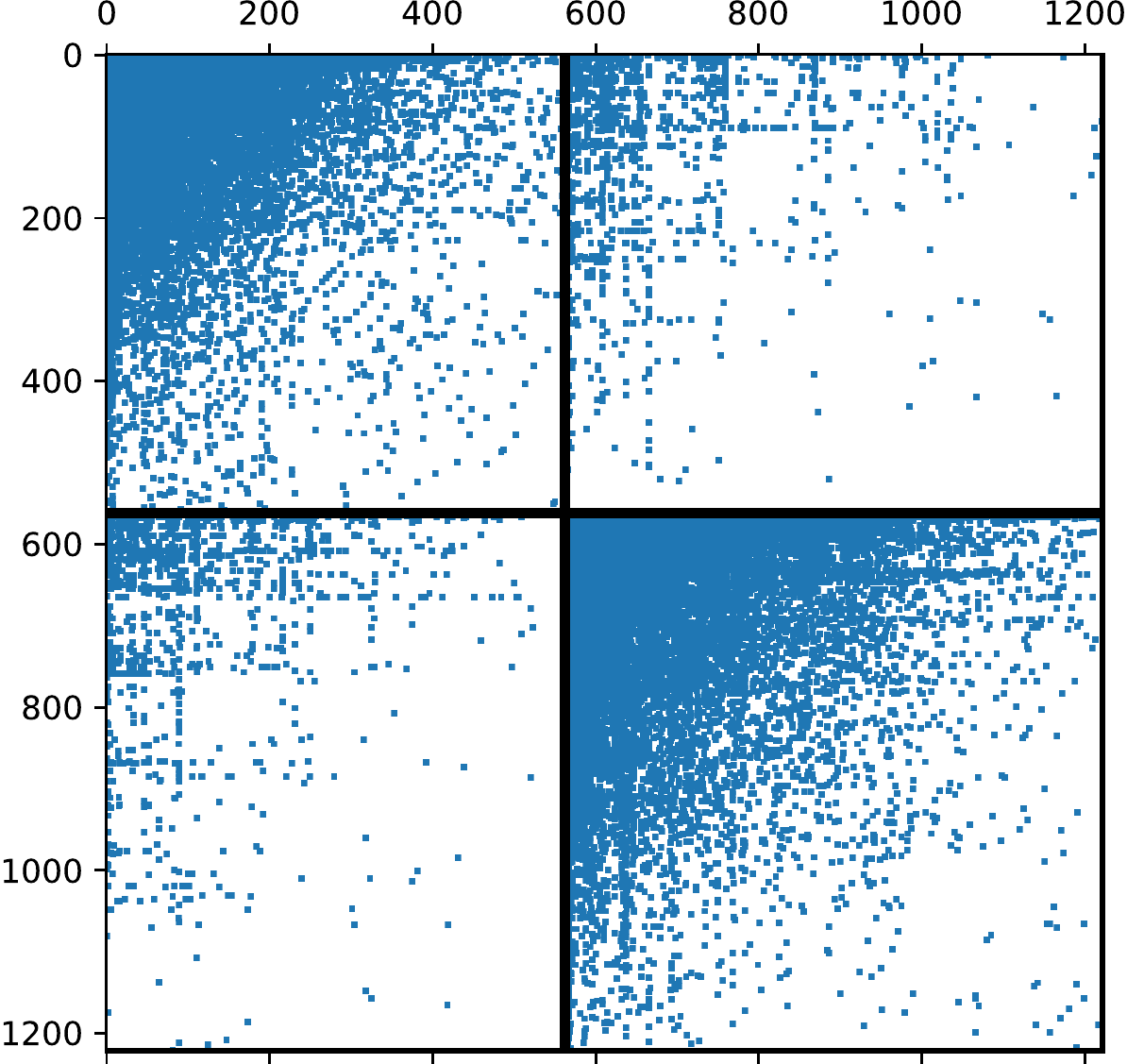}
\includegraphics[width=0.19\linewidth]{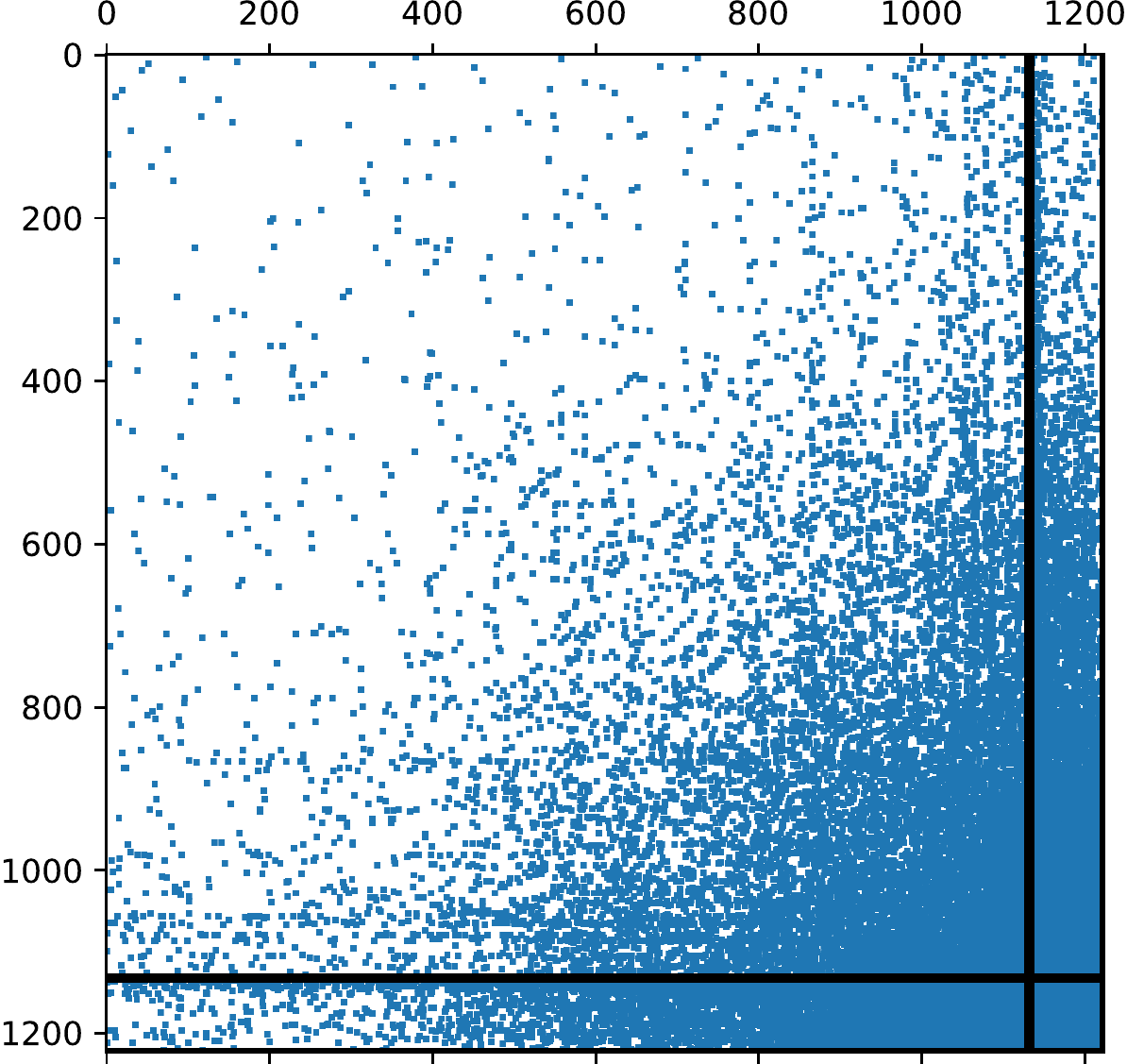}
\newline
\subfigure[Ground truth]{\includegraphics[width=0.19\linewidth]{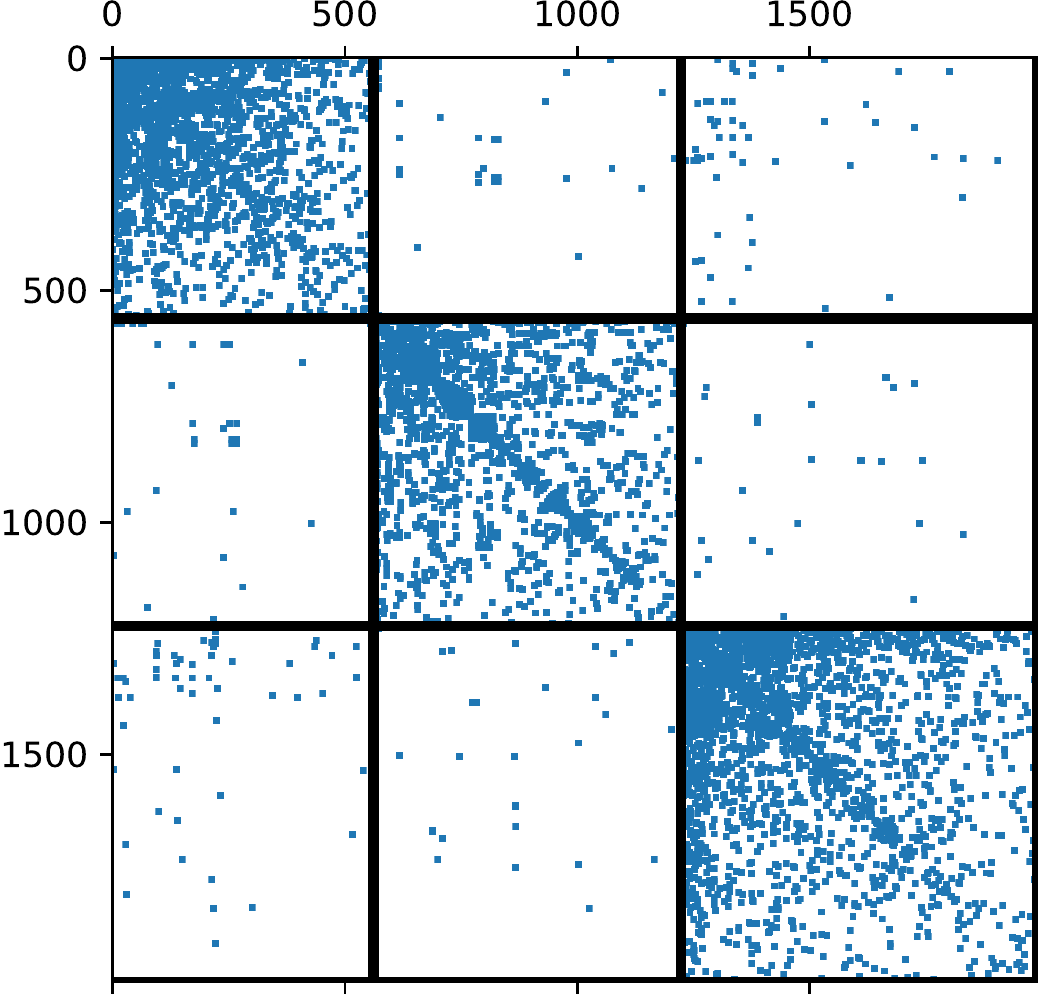}}
\subfigure[IG-NR]{\includegraphics[width=0.19\linewidth]{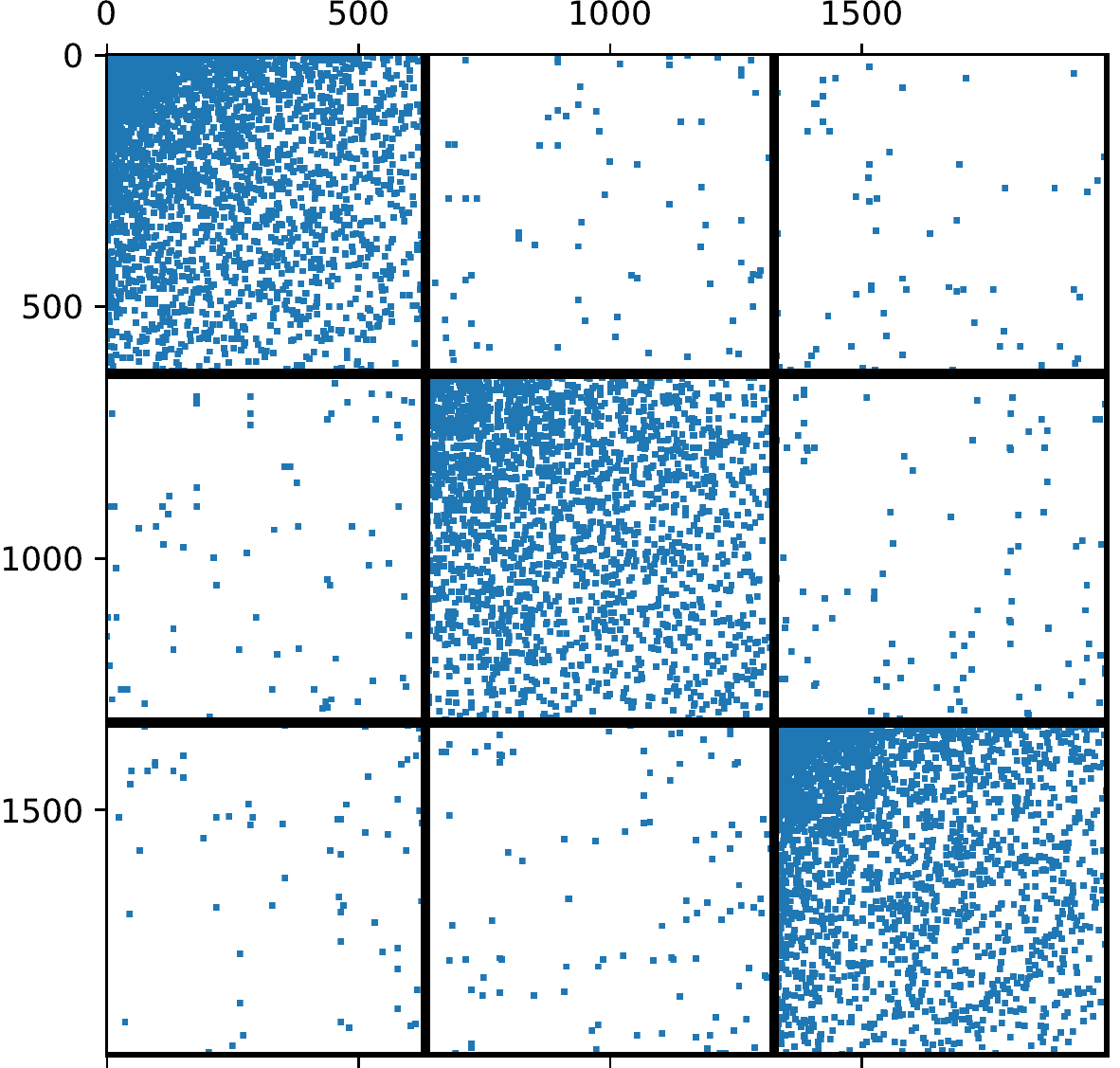}}
\subfigure[GIG-NR]{\includegraphics[width=0.19\linewidth]{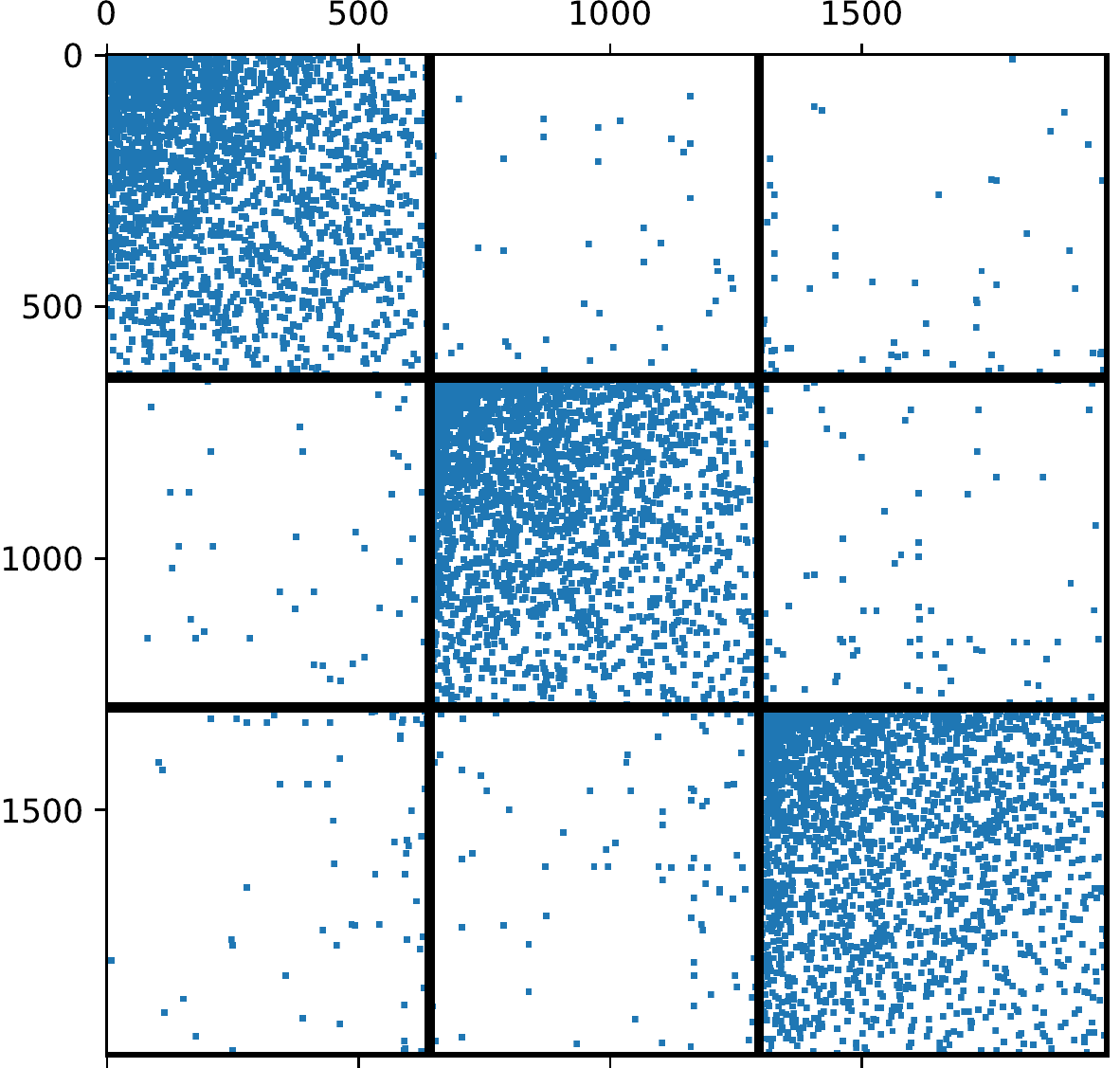}}
\subfigure[CGGP]{\includegraphics[width=0.19\linewidth]{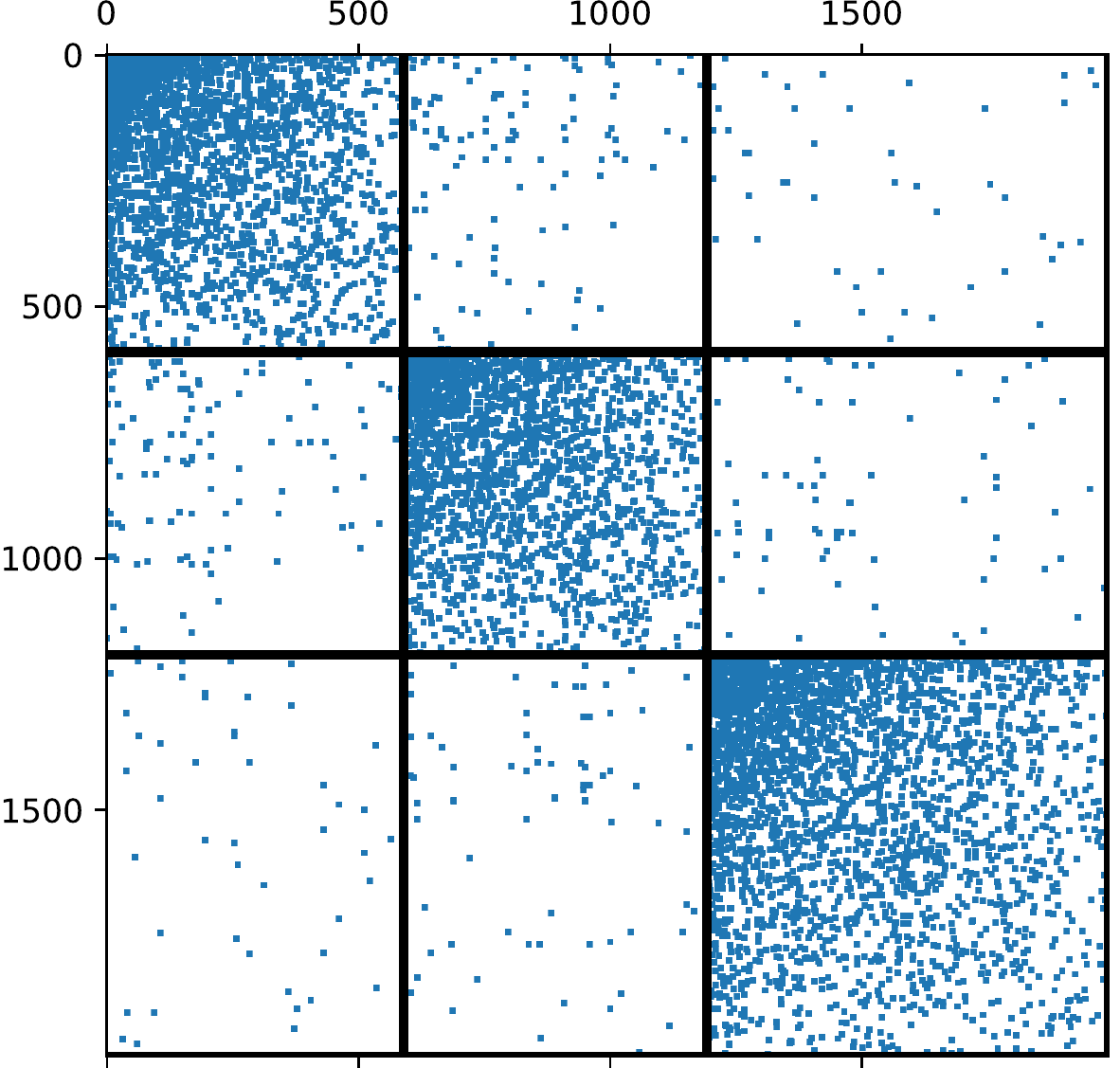}}
\subfigure[MMSB]{\includegraphics[width=0.19\linewidth]{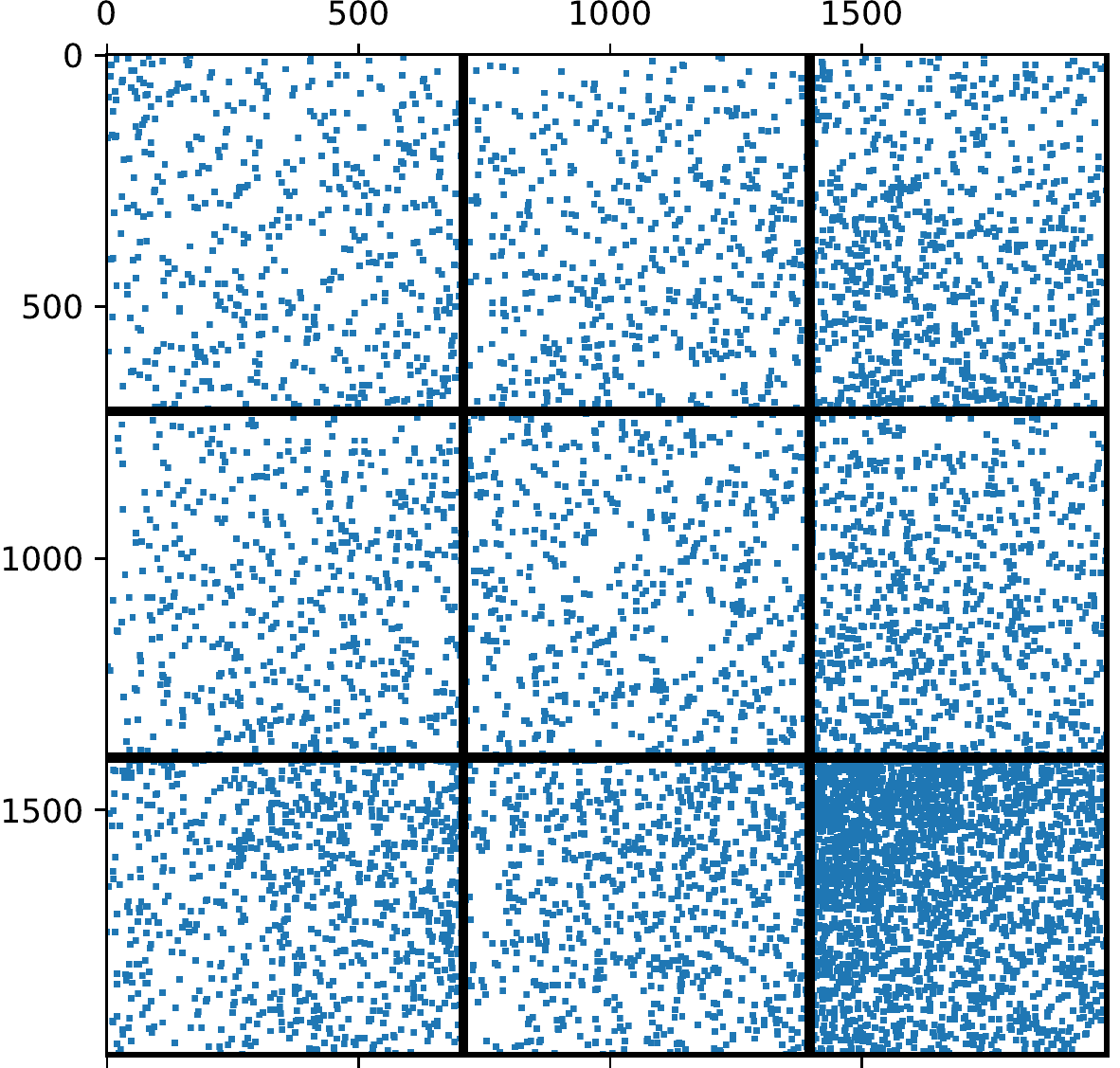}}
\caption{Group truth and latent communities discovered by the random graph models for (top row) \texttt{polblogs} and (bottom row) \texttt{DBLP}. 
 The nodes within communities are sorted according to their degree for the ground truth, and according to their estimated node popularity parameter for the different models.}
\label{fig:discovered_communities}
\end{figure}

We present the community structures discoverd by IG, GIG, CGGP and MMSB in \cref{fig:discovered_communities}. IG, GIG and CGGP discovered reasonable communities
where the edge densities within communities are much higher than the edge densities between communities. However, MMSB completely failed to discover the communities. The fact that the models without degree heterogeneity fail to capture community structures has been reported in various works~\citep{Karrer2011,Gopalan2013, Todeschini2016}, and our results confirm it.

\end{appendices}

\end{document}